\documentclass[11pt,letterpaper]{article}

\usepackage{iccv}
\usepackage{times}
\usepackage{epsfig}
\usepackage{graphicx}
\usepackage{amsmath}
\usepackage{amssymb}

\usepackage{bm}
\usepackage{subfigure}
\usepackage{color}
\usepackage{multirow}
\usepackage[table]{xcolor}
\usepackage{colortbl}
\usepackage{adjustbox}

\usepackage[pagebackref=true,breaklinks=true,letterpaper=true,colorlinks,bookmarks=false]{hyperref}

\iccvfinalcopy % *** Uncomment this line for the final submission

\def\iccvPaperID{507} % *** Enter the ICCV Paper ID here

\begin{document}

%%%%%%%%% TITLE
\title{Toward Real-World Single Image Super-Resolution: \\A New Benchmark and A New Model}

\author{Jianrui Cai$^{1,}\footnotemark[1]$~~, 
Hui Zeng$^{1, }\footnotemark[1]$~~, 
Hongwei Yong$^{1}$, Zisheng Cao$^{2}$, Lei Zhang$^{1,3,}\footnotemark[4]$\\
$^{1}$The Hong Kong Polytechnic University,
$^{2}$DJI Co.,Ltd,
$^{3}$DAMO Academy, Alibaba Group\\
{\tt\small \{csjcai, cshzeng, cshyong, cslzhang\}@comp.polyu.edu.hk, zisheng.cao@dji.com}
}

\maketitle
%\thispagestyle{empty}

%%%%%%%%% ABSTRACT
\begin{abstract}
Most of the existing learning-based single image super-resolution (SISR) methods are trained and evaluated on simulated datasets, where the low-resolution (LR) images are generated by applying a simple and uniform degradation (i.e., bicubic downsampling) to their high-resolution (HR) counterparts. 
However, the degradations in real-world LR images are far more complicated. 
As a consequence, the SISR models trained on simulated data become less effective when applied to practical scenarios. 
In this paper, we build a real-world super-resolution (RealSR) dataset where paired LR-HR images on the same scene are captured by adjusting the focal length of a digital camera. 
An image registration algorithm is developed to progressively align the image pairs at different resolutions. 
Considering that the degradation kernels are naturally non-uniform in our dataset, we present a Laplacian pyramid based kernel prediction network (LP-KPN), which efficiently learns per-pixel kernels to recover the HR image. 
Our extensive experiments demonstrate that SISR models trained on our RealSR dataset deliver better visual quality with sharper edges and finer textures on real-world scenes than those trained on simulated datasets. 
Though our RealSR dataset is built by using only two cameras (Canon 5D3 and Nikon D810), the trained model generalizes well to other camera devices such as Sony a7II and mobile phones.
\end{abstract}

\renewcommand{\thefootnote}{\fnsymbol{footnote}}
\footnotetext[1]{The first two authors contribute equally to this work. A part of this dataset was used in the image super-resolution challenge in the NTIRE 2019 challenges \cite{NTIRE2019}.}
\footnotetext[4]{Corresponding author: Lei Zhang}
%%%%%%%%% BODY TEXT
\section{Introduction}
\begin{figure}
\centering
\footnotesize
\subfigure{
\begin{minipage}{0.4\linewidth}
\centering
\includegraphics[width=1\textwidth]{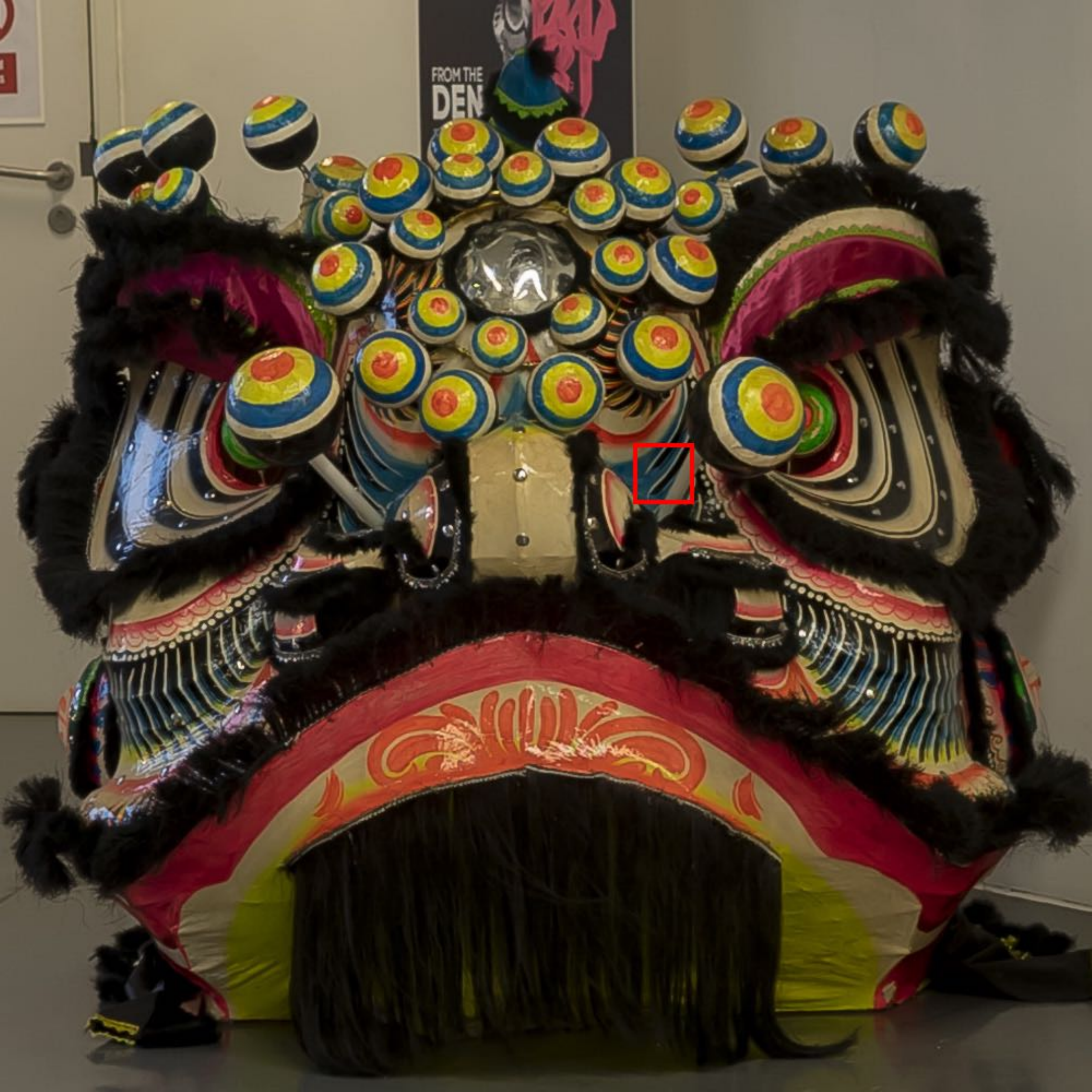}
{(a) Image captured by Sony a7II}
\end{minipage}
}
\footnotesize
\subfigure{
\begin{minipage}{0.1899\linewidth}
\centering
\includegraphics[width=0.985\textwidth]{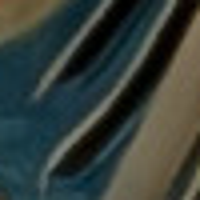}
{(b) Bicubic}
\centering
\includegraphics[width=0.985\textwidth]{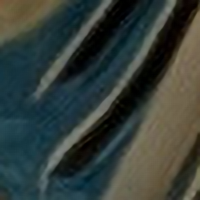}
{(c) RCAN + BD}
\end{minipage}
}\\
\footnotesize
\subfigure{
\begin{minipage}{0.193\linewidth}
\centering
\includegraphics[width=0.985\textwidth]{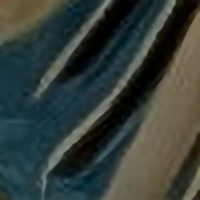}
{(d) RCAN + MD}
\end{minipage}
}
\footnotesize
\subfigure{
\begin{minipage}{0.193\linewidth}
\centering
\includegraphics[width=0.985\textwidth]{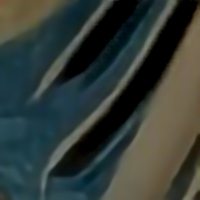}
{(e) RCAN + RealSR}
\end{minipage}
}
\footnotesize
\subfigure{
\begin{minipage}{0.193\linewidth}
\centering
\includegraphics[width=0.985\textwidth]{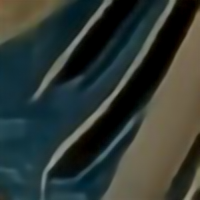}
{(f) LP-KPN + RealSR}
\end{minipage}
}
\caption{
The SISR results ($\times4$) of (a) a real-world image captured by a Sony a7II camera. 
SISR results generated by (b) bicubic interpolator, RCAN models \cite{zhang2018image} trained using image pairs (in DIV2K \cite{timofte2017ntire}) with (c) bicubic degradation (BD), (d) multiple simulated degradations (MD) \cite{zhang2018learning}, and (e) authentic distortions in our RealSR dataset. 
(f) SISR result by the proposed LP-KPN model trained on our dataset. 
Note that our RealSR dataset is collected by Canon 5D3 and Nikon D810 cameras.
}
\label{figure:SampleIntro}
\end{figure}

Single image super-resolution (SISR) \cite{glasner2009super} aims to recover a high-resolution (HR) image from its low-resolution (LR) observation. 
SISR has been an active research topic for decades \cite{park2003super,yang2018deep, timofte2017ntire, timofte2018ntire, blau20182018} because of its high practical values in enhancing image details and textures. 
Since SISR is a severely ill-posed inverse problem, learning image prior information from the HR and/or LR exemplar images \cite{glasner2009super, freedman2011image, yang2013fast, huang2015single, freeman2000learning, chang2004super, kim2010single,yang2010image, dong2011image, jia2013image, timofte2013anchored, schulter2015fast} plays an indispensable role in recovering the details from an LR input image. 
Benefitting from the rapid development of deep convolutional neural networks (CNNs) \cite{lecun2015deep}, recent years have witnessed an explosive spread of training CNN models to perform SISR, and the performance has been consistently improved by designing new CNN architectures \cite{dong2014learning, wang2015deep,shi2016real, kim2016accurate, tai2017image, lim2017enhanced, zhang2018residual, zhang2018image} and loss functions \cite{johnson2016perceptual, ledig2017photo, sajjadi2017enhancenet}.
Though significant advances have been made, most of the existing SISR methods are trained and evaluated on simulated datasets which assume simple and uniform degradation (\ie, bicubic degradation). 
Unfortunately, SISR models trained on such simulated datasets are hard to generalize to practical applications since the authentic degradations in real-world LR images are much more complex \cite{yang2014single, kohler2018bridging}.
Fig. \ref{figure:SampleIntro} shows the SISR results of a real-world image captured by a Sony a7II camera. 
We utilize the state-of-the-art RCAN method \cite{zhang2018image} to train three SISR models using simulated image pairs (in DIV2K \cite{timofte2017ntire}) with bicubic degradation, multiple simulated degradations \cite{zhang2018learning} and image pairs with authentic distortions in our dataset to be constructed in this paper. 
The results clearly show that, compared with the simple bicubic interpolator (Fig. \ref{figure:SampleIntro}(b)), the RCAN models trained on simulated datasets (Figs. \ref{figure:SampleIntro}(c)$\sim$\ref{figure:SampleIntro}(d)) do not show clear advantages on real-world images. 
It is thus highly desired that we can have a training dataset consisting of real-world, instead of simulated, LR and HR image pairs. 
However, constructing such a real-world super-resolution (RealSR) dataset is a non-trivial job since the ground-truth HR images are very difficult to obtain. 
To the best of our knowledge, only two recent attempts have been made in the laboratory environment, where complicated devices were used to collect image pairs on very limited scenes \cite{Qu2016CapturingGT, kohler2018bridging}. 
In this work, we aim to construct a more general and practical RealSR dataset using a flexible and easy-to-reproduce method. 
Specifically, we capture images of the same scene using fixed digital single-lens reflex (DSLR) cameras with different focal lengths. 
By increasing the focal length, finer details of the scene can be naturally recorded into the camera sensor. 
In this way, HR and LR image pairs on different scales can be collected. 
However, in addition to the change of field of view (FoV), adjusting focal length can result in many other changes in the imaging process, such as shift of optical center, variation of scaling factors, different exposure time and lens distortion. 
We thus develop an effective image registration algorithm to progressively align the image pairs such that the end-to-end training of SISR models can be performed. 
The constructed RealSR dataset contains various indoor and outdoor scenes taken by two DSLR cameras (Canon 5D3 and Nikon D810), providing a good benchmark for training and evaluating SISR algorithms in practical applications.
Compared with the previous simulated datasets, the image degradation process in our RealSR dataset is much more complicated. 
In particular, the degradation is spatially variant since the blur kernel varies with the depth of content in a scene. 
This motivates us to train a kernel prediction network (KPN) for the real-world SISR task. 
The idea of kernel prediction is to explicitly learn a restoration kernel for each pixel, and it has been employed in applications such as denoising \cite{bako2017kernel, mildenhall2018burst,vogels2018denoising}, dynamic deblurring \cite{sun2015learning, gong2017motion} and video interpolation \cite{niklaus2017video, niklaus2017}. 
Though effective, the memory and computational cost of KPN is quadratically increased with the kernel size. 
To obtain as competitive SISR performance as using large kernel size while achieving high computational efficiency, we propose a Laplacian pyramid based KPN (LP-KPN) which learns per-pixel kernels for the decomposed image pyramid. 
Our LP-KPN can leverage rich information using a small kernel size, leading to effective and efficient real-world SISR performance. 
Figs. \ref{figure:SampleIntro}(e) and \ref{figure:SampleIntro}(f) show the SISR results of RCAN \cite{zhang2018image} and LP-KPN models trained on our RealSR dataset, respectively. 
One can see that both of them deliver much better results than the RCAN models trained on simulated data, while our LP-KPN ($46$ conv layers) can output more distinct result than RCAN (over 400 conv layers) using much fewer layers.
The contributions of this work are twofold:
\begin{itemize}
\item 
We build a RealSR dataset consisting of HR and LR image pairs with different scaling factors. 
It provides, to the best of our knowledge, the first general purpose benchmark for real-world SISR model training and evaluation. 
\item 
We present an LP-KPN model and validate its efficiency and effectiveness in real-world SISR.  
\end{itemize}
Extensive experiments are conducted to quantitatively and qualitatively analyze the performance of our RealSR dataset in training SISR models. 
Though the dataset in its current version is built using only two cameras, the trained SISR models exhibit good generalization capability to images captured by other types of camera devices.
\section{Related Work}
\noindent \textbf{SISR datasets.}
There are several popular datasets, including Set5 \cite{bevilacqua2012low}, Set14 \cite{zeyde2010single}, BSD300 \cite{martin2001database}, Urban100 \cite{huang2015single}, Manga109 \cite{matsui2017sketch} and DIV2K \cite{timofte2017ntire} that have been widely used for training and evaluating the SISR methods. 
In all these datasets, the LR images are generally synthesized by a simple and uniform degradation process such as bicubic downsampling or Gaussian blurring followed by direct downsampling \cite{dong2013nonlocally}. 
The SISR Models trained on these simulated data may exhibit poor performance when applied to real LR images where the degradation deviates from the simulated ones \cite{efrat2013accurate}. 
To improve the generalization capability, Zhang \etal \cite{zhang2018learning} trained their model using multiple simulated degradations and Bulat \etal \cite{bulat2018learn} used a GAN \cite{goodfellow2014generative} to generate the degradation process. 
Although these more advanced methods can simulate more complex degradation, there is no guarantee that such simulated degradation can approximate the authentic degradation in practical scenarios which is usually very complicated \cite{kohler2018bridging}. 
It is very challenging to get the ground-truth HR image for an LR image in real-world scenarios, making the training and evaluation of real-world SISR models difficult. 
To the best of our knowledge, only two recent attempts have been made on capturing real-world image pairs for SISR. 
Qu \etal \cite{Qu2016CapturingGT} put two cameras together with a beam splitter to collect a dataset with paired face images. 
K{\"o}hler \etal \cite{kohler2018bridging} employed hardware binning on the sensor to capture LR images and used multiple postprocessing steps to generate different versions of an LR image.
However, both the datasets were collected in indoor laboratory environment and very limited number of scenes (31 face images in \cite{Qu2016CapturingGT} and 14 scenes in \cite{kohler2018bridging}) were included. 
Different from them, our dataset is constructed by adjusting the focal length of DSLR cameras which naturally results in image pairs at different resolutions, and it contains $234$ scenes in both indoor and outdoor environments.
\noindent \textbf{Kernel prediction networks.}
Considering that the degradation kernel in our RealSR dataset is spatially variant, we propose to train a kernel prediction network (KPN) for real-world SISR. 
The idea of KPN was first proposed in \cite{bako2017kernel} to denoise Monte Carlo renderings and it has proven to have faster convergence and better stability than direct prediction \cite{vogels2018denoising}. 
Mildenhall \etal \cite{mildenhall2018burst} trained a KPN model for burst denoising and obtained state-of-the-art performance on both synthetic and real data.
Similar ideas have been employed in estimating the blur kernels in dynamic deblurring \cite{sun2015learning, gong2017motion} or convolutional kernels in video interpolation \cite{niklaus2017video, niklaus2017}. 
We are among the first to train a KPN for SISR and we propose the LP-KPN to perform kernel prediction in the scale space with high efficiency.
\section{Real-world SISR Dataset}\label{dataset}

To build a dataset for learning and evaluating real-world SISR models, we propose to collect images of the same scene by adjusting the lens of DSLR cameras. 
Sophisticated image registration operations are then performed to generate the HR and LR pairs of the same content. 
The detailed dataset construction process is presented in this section.  
\begin{figure}[t]
\centering
\subfigure{
\begin{minipage}[b]{0.62\linewidth}
\centering
\includegraphics[width=1\textwidth]{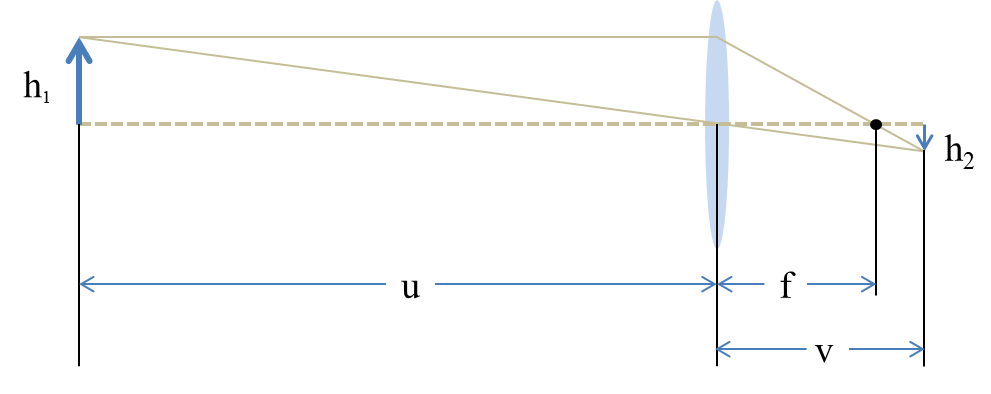}
\end{minipage}}
\caption{Illustration of thin lens. $u, v, f$ represent the object distance, image distance and focal length, respectively. $h_{1}$ and $h_{2}$ denote the size of object and image.}
\label{figure:thinlens}
\end{figure}
\subsection{Image formation by thin lens}
The DSLR camera imaging system can be approximated as a thin lens \cite{wiki:lens}. 
An illustration of the image formation process by thin lens is shown in Fig. \ref{figure:thinlens}.
We denote the object distance, image distance and focal length by $u, v, f$, and denote the size of object and image by $h_{1}$ and $h_{2}$, respectively. 
The lens equation is defined as follows \cite{wiki:lens}:
\begin{equation}\label{equ:lensequ}
\frac{1}{f} = \frac{1}{u} + \frac{1}{v}.
\end{equation}
The magnification factor $M$ is defined as the ratio of the image size to the object size:
\begin{equation}\label{equ:magnification1}
M = \frac{h_2}{h_1} = \frac{v}{u}.
\end{equation}
In our case, the static images are taken at a distance (\ie, $u$) larger than 3.0m. Both $h_1$ and $u$ are fixed and $u$ is much larger than $f$ (the largest $f$ is 105mm). 
Combining Eq. (\ref{equ:lensequ}) and Eq. (\ref{equ:magnification1}), and considering the fact that $u\gg f$, we have:
\begin{equation}\label{equ:magnification2}
h_2 = \frac{f}{u-f}h_1 \approx \frac{f}{u}h_1.
\end{equation}
Therefore, $h_2$ is approximately linear to $f$. 
By increasing the focal length $f$, larger images with finer details will be recorded in the camera sensor. 
The scaling factor can also be controlled (in theory) by choosing specific values of $f$. 
\subsection{Data collection}
\begin{table}[t]
\small
\caption{Number of image pairs for each camera at each scaling factor.}
\label{table:imagenum}
\begin{center}
\begin{tabular}{|c|c|c|c||c|c|c|}
\hline
			     Camera               &\multicolumn{3}{c||}{Canon 5D3} 	   &\multicolumn{3}{c|}{Nikon D810}\\
\hline
         	     Scale                & $\times$2 & $\times$3 & $\times$4  & $\times$2 & $\times$3 & $\times$4\\
\hline \hline
\rowcolor{gray!40} \# image pairs     & 86 &  117 &  86 				   & 97 &  117 &  92 \\
\hline
\end{tabular}
\end{center}
\end{table}

We used two full frame DSLR cameras (Canon 5D3 and Nikon D810) to capture images for data collection. 
The resolution of Canon 5D3 is $5760\times3840$, and that of Nikon D810 is $7360\times4912$. 
To cover the common scaling factors (\eg, $\times 2, \times 3, \times 4$) used in most previous SISR datasets, both cameras were equipped with one 24$\sim$105mm, $f$/4.0 zoom lens. 
For each scene, we took photos using four focal lengths: 105mm, 50mm, 35mm, and 28mm. 
Images taken by the largest focal length are used to generate the ground-truth HR images, and images taken by the other three focal lengths are used to generate the LR versions. 
We choose 28mm rather than 24mm because lens distortion at 24mm is more difficult to correct in post-processing, which results in less satisfied quality in image pair registration.
The camera was set to aperture priority mode and the aperture was adjusted according to the depth-of-field (DoF) \cite{wiki:DoF}. 
Basically, the selected aperture value should make the DoF large enough to cover the scene and avoid severe diffraction. 
Small ISO is preferred to alleviate noise. 
The focus, white balance, and exposure were set to automatic mode. 
The center-weighted metering option was selected since only the center region of captured images were used in our final dataset. 
For stabilization, the camera was fixed on a tripod and a bluetooth remote controller was used to control the shutter. 
Besides, lens stabilization was turned off and the reflector was pre-rised when taking photos.
To ensure the generality of our dataset, we took photos in both indoor and outdoor environment. 
Scenes with abundant texture are preferred considering that the main purpose of super-resolution is to recover or enhance image details. 
For each scene, we first captured the image at 105mm focal length and then manually decreased the focal length to take three down-scaled versions. 
$234$ scenes were captured, and there are no overlapped scenes between the two cameras.
After discarding images having moving objects, inappropriate exposure, and blur, we have $595$ HR and LR image pairs in total. 
The numbers of image pairs for each camera at each scaling factor are listed in Table \ref{table:imagenum}. 
\subsection{Image pair registration}\label{sec:reg}
\begin{figure}[t]
\centering
\subfigure{
\begin{minipage}[b]{0.8\linewidth}
\centering
\includegraphics[width=1\textwidth]{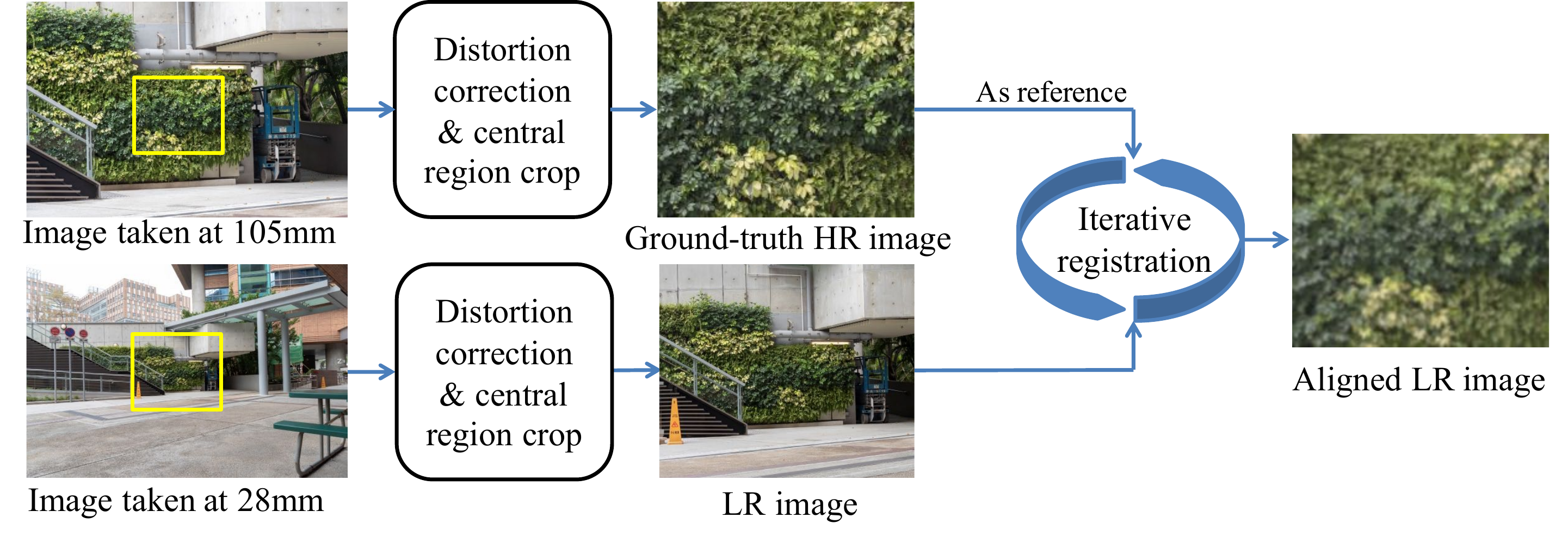}
\end{minipage}}
\caption{Illustration of our image pair registration process.}
\label{figure:registration}
\end{figure}

Although it is easy to collect images on different scales by zooming the lens of a DSLR camera, it is difficult to obtain pixel-wise aligned image pairs because the zooming of lens brings many uncontrollable changes. 
Specifically, images taken at different focal lengths suffer from different lens distortions and usually have different exposures. 
Moreover, the optical center will also shift when zooming the focal length because of the inherent defect of lens. 
Even the scaling factors are varying slightly because the lens equation (Eq. (\ref{equ:lensequ})) cannot be precisely satisfied in practical focusing process. 
With the above factors, none of the existing image registration algorithms can be directly used to obtain accurate pixel-wise registration of two images captured under different focal length. 
We thus develop an image registration algorithm to progressively align such image pairs to build our RealSR dataset. 
The registration process is illustrated in Fig. \ref{figure:registration}. 
We first import the images with meta information into PhotoShop to correct the lens distortion. 
However, this step cannot perfectly correct the lens distortion especially for the region distant from the optical center. 
We thus further crop the interested region around the center of the image, where distortion is not severe and can be well corrected. 
The cropped region from the image taken at 105mm focal length is used as the ground-truth HR image, whose LR counterparts are to be registered from the original images taken at 50mm, 35mm, or 28mm focal length. 
Since there is certain luminance and scale difference between images taken at different focal lengths, those popular keypoint based image registration algorithms such as SURF \cite{bay2006surf} and SIFT \cite{lowe2004distinctive} cannot always achieve pixel-wise registration, which is necessary for our dataset. 
To obtain accurate image pair registration, we develop a pixel-wise registration algorithm which simultaneously considers luminance adjustment. 
Denote by $\mathbf{I}_H$ and $\mathbf{I}_L$ the HR image and the LR image to be registered, our algorithm minimizes the following objective function:
\begin{equation}
\min_{\boldsymbol{\tau}}||\alpha C(\boldsymbol{\tau}\circ\mathbf{I}_L)+\beta-\mathbf{I}_H||_p^p,
\end{equation}
where $\boldsymbol{\tau}$ is an affine transformation matrix, $C$ is a cropping operation which makes the transformed $\mathbf{I}_L$ have the same size as $\mathbf{I}_H$, $\alpha$ and $\beta$ are luminance adjustment parameters, $||\cdot||_p $ is a robust $L_p$-norm $(p\leq1)$, \eg, $L_1$-norm.
The above objective function is solved in an iterative manner. 
At the beginning, according to Eq. (\ref{equ:magnification2}), the $\boldsymbol{\tau}$ is initialized as a scaling transformation with scaling factor calculated as the ratio of two focal lengths. 
Let $\mathbf{I}_L'=C(\boldsymbol{\tau}\circ\mathbf{I}_L)$. 
With $\mathbf{I}_L'$ and $\mathbf{I}_H$ fixed, the parameters for luminance adjustment can be obtained by $\alpha=\text{std}(\mathbf{I}_H)/\text{std}(\mathbf{I}_L')$ and $\beta=\text{mean}(\mathbf{I}_H)-\alpha \text{mean}(\mathbf{I}_L')$, which can ensure $\mathbf{I}_L'$ having the same pixel mean and variance as $\mathbf{I}_H$ after luminance adjustment.
Then we solve the affine transformation matrix $\boldsymbol{\tau}$ with $\alpha$ and $\beta$ fixed.
According to \cite{odobez1995robust,yong2018robust}, the objective function w.r.t. $\boldsymbol{\tau}$ is nonlinear, which can be iteratively solved by a locally linear approximation:
\begin{equation}
\min_{\Delta\boldsymbol{\tau}}||\alpha C(\boldsymbol{\tau}\circ\mathbf{I}_L)+\beta+\alpha\mathbf{J}\Delta\boldsymbol{\tau}-\mathbf{I}_H||_p^p,
\end{equation}
where $\mathbf{J}$ is the Jacobian matrix of $C(\boldsymbol{\tau}\circ\mathbf{I}_L)$ w.r.t. $\boldsymbol{\tau}$, and this objective function can be solved by an iteratively reweighted least square problem (IRLS) as follows \cite{chartrand2008iteratively}:
\begin{equation}
\min_{\Delta\boldsymbol{\tau}}||\mathbf{w}\odot(\mathbf{A}\Delta\boldsymbol{\tau}-\mathbf{b})||_2^2,
\end{equation}
where $\mathbf{A}=\alpha\mathbf{J}$, $\mathbf{b}=\mathbf{I}_H-(\alpha C(\boldsymbol{\tau}\circ\mathbf{I}_L)+\beta)$, $\mathbf{w}$ is the weight matrix and $\odot$ denotes element-wise multiplication. 
Then we can obtain:
\begin{equation}
\Delta\boldsymbol{\tau}=(\mathbf{A}'\text{diag}(\mathbf{w})^2\mathbf{A})^{-1}\mathbf{A}'\text{diag}(\mathbf{w})^2\mathbf{b},
\end{equation}
and $\boldsymbol{\tau}$ can be updated by: $\boldsymbol{\tau}=\boldsymbol{\tau}+\Delta\boldsymbol{\tau}$.
We iteratively estimate the luminance adjustment parameters and the affine transformation matrix. 
The optimization process converges within $5$ iterations since our prior information of the scaling factor provides a good initialization of $\boldsymbol{\tau}$. 
After convergence, we can obtain the aligned LR image as $\textbf{I}_{L}^A = \alpha C(\boldsymbol{\tau}\circ\mathbf{I}_L)+\beta$.
\begin{figure*}[t]
\centering
\subfigure{
\begin{minipage}[b]{1.0\linewidth}
\centering
\includegraphics[width=1\textwidth]{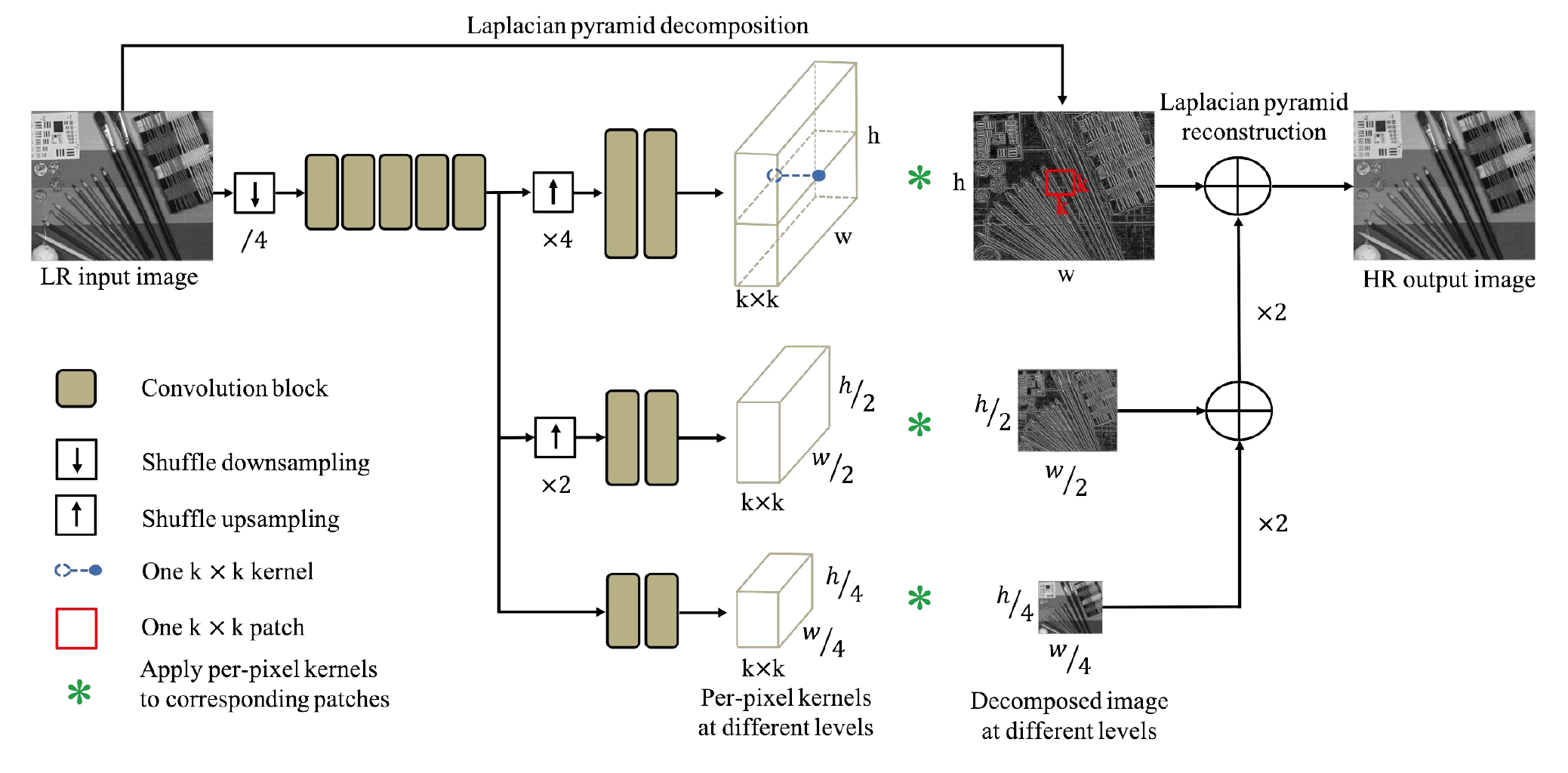}
\end{minipage}}
\caption{Framework of the Laplacian pyramid based kernel prediction network. By decomposing the image into a Laplacian pyramid, using small kernels can leverage rich neighborhood information for super-resolution.}
\label{figure:framework}
\end{figure*}
\section{Laplacian Pyramid based Kernel Prediction Network}
In Section \ref{dataset}, we have constructed a new real-world super-resolution (RealSR) dataset, which consists of pixel-wise aligned HR and LR image pairs $\{\textbf{I}_{H}, \textbf{I}_{L}^A\}$ of size $h\times w$.
Now the problem turns to how to learn an effective network to enhance $\textbf{I}_{L}^A$ to $\textbf{I}_{H}$. 
For LR images in our RealSR dataset, the blur kernel varies with the depth in a scene \cite{wiki:Circleofconfusion} and the DoF \cite{wiki:DoF} changes with the focal length. 
Training an SISR model which directly transforms the LR image to the HR image, as done in most of the previous CNN based SISR methods, may not be the cost-effective way. 
We therefore propose to train a kernel prediction network (KPN) which explicitly learns an individual kernel for each pixel. 
Compared with those direct pixel synthesis networks, KPN has proven to have advantages in efficiency, interpretability and generalization capability in tasks of denoising, dynamic deblurring, etc., \cite{bako2017kernel, mildenhall2018burst, vogels2018denoising, sun2015learning, gong2017motion, kong2018image}.
The KPN takes the $\textbf{I}_{L}^A$ as input and outputs a kernel tensor $\mathbf{T}\in R^{(k\times k)\times h\times w}$, in which each vector in channel dimension $\mathbf{T}(i,j)\in R^{(k\times k)}$ can be reshaped into a $k\times k$ kernel $\mathbf{K}(i,j)$. The reshaped per-pixel kernel $\mathbf{K}(i,j)$ is applied to the $k\times k$ neighborhood of each pixel in the input LR image $\textbf{I}_{L}^A(i,j)$ to reproduce the HR output. The predicted HR image, denoted by $\textbf{I}_{H}^P$, is obtained by:
\begin{equation}\label{equ:kpn}
\textbf{I}_{H}^P(i,j) = \langle \mathbf{K}(i,j),V(\textbf{I}_{L}^A(i,j))\rangle,
\end{equation}
where $V(\textbf{I}_{L}^A(i,j))$ represents a $k\times k$ neighborhood of pixel $\textbf{I}_{L}^A(i,j)$ and $\langle\bm{\cdot}\rangle$ denotes the inner product operation.
Eq. (\ref{equ:kpn}) shows that the output pixel is a weighted linear combination of the neighboring pixels in the input image. 
To obtain good performance, a large kernel size is necessary to leverage richer neighborhood information, especially when only a single frame image is used. 
On the other hand, the predicted kernel tensor $\mathbf{T}$ grows quadratically with the kernel size $k$, which can result in high computational and memory cost in practical applications. 
In order to train a both effective and efficient KPN, we propose a Laplacian pyramid based KPN (LP-KPN).
The framework of our LP-KPN is shown in Fig. \ref{figure:framework}. 
As in many SR methods \cite{lim2017enhanced, timofte2018ntire}, our model works on the Y channel of YCbCr space. 
The Laplacian pyramid decomposes an image into several levels of sub-images with downsampled resolution and the decomposed images can exactly reconstruct the original image. 
Using this property, the Y channel of an LR input image $\textbf{I}_{L}^A$ is decomposed into a three-level image pyramid \{$\mathbf{S}_0,\mathbf{S}_1,\mathbf{S}_2$\}, where $\mathbf{S}_0\in R^{h\times w}$, $\mathbf{S}_1\in R^{\frac{h}{2}\times \frac{w}{2}}$, and $\mathbf{S}_2\in R^{\frac{h}{4}\times \frac{w}{4}}$. 
Our LP-KPN takes the LR image as input and predicts three kernel tensors \{$\mathbf{T}_0,\mathbf{T}_1,\mathbf{T}_2$\} for the image pyramid, where $\mathbf{T}_0\in R^{(k\times k)\times h\times w}$, $\mathbf{T}_1\in R^{(k\times k)\times \frac{h}{2}\times \frac{w}{2}}$, and $\mathbf{T}_2\in R^{(k\times k)\times \frac{h}{4}\times \frac{w}{4}}$. 
The learned kernel tensors \{$\mathbf{T}_0,\mathbf{T}_1,\mathbf{T}_2$\} are applied to the corresponding image pyramid \{$\mathbf{S}_0,\mathbf{S}_1,\mathbf{S}_2$\}, using the operation in Eq. (\ref{equ:kpn}), to restore the Laplacian decomposition of HR image at each level. 
Finally, the Laplacian pyramid reconstruction is conducted to obtain the HR image. 
Benefitting from the Laplacian pyramid, learning three $k\times k$ kernels can equally lead to a receptive field with size $4k\times4k$ at the original resolution, which significantly reduces the computational cost compared to directly learning one $4k\times4k$ kernel.
The backbone of our LP-KPN consists of $17$ residual blocks, with each residual block containing $2$ convolutional layers and a ReLU function (similar structure to \cite{lim2017enhanced}). 
To improve the efficiency, we shuffle \cite{shi2016real} the input LR image with factor $\frac{1}{4}$ (namely, the $h\times w$ image is shuffled to $16$ $\frac{h}{4}\times\frac{w}{4}$ images) and input the shuffled images to the network.
Most convolutional blocks are shared by the three levels of kernels except for the last few layers. 
One $\times4$ and one $\times2$ shuffle operation are performed to upsample the spatial resolution of the latent image representations at two lower levels, followed by individual convolutional blocks. 
Our LP-KPN has a total of $46$ convolutional layers, which is much less than the previous state-of-the-art SISR models \cite{lim2017enhanced, zhang2018residual, zhang2018image}. 
The detailed network architecture can be found in the $\textbf{supplementary material}$. 
The $L_{2}$-norm loss function $\mathcal{L}(\mathbf{I}_H,\textbf{I}_H^P) = ||\textbf{I}_{H}-\textbf{I}_{H}^P||_2^2$ is employed to minimize the pixel-wise distance between the model prediction $\textbf{I}_{H}^P$ and the ground-truth HR image $\textbf{I}_H$.
\section{Experiments}
\noindent \textbf{Experimental setup.}
The number of image pairs in our RealSR dataset is reported in Table \ref{table:imagenum}. 
We randomly selected $15$ image pairs at each scaling factor for each camera to form the testing set, while using the remaining image pairs as training set. 
Except for cross-camera testing, images from both the Canon and Nikon cameras were combined for training and testing. 
Following the previous work \cite{lim2017enhanced, zhang2018image, timofte2018ntire}, the SISR results were evaluated using PSNR and SSIM \cite{wang2004image} indices on the Y channel in the YCbCr space. 
The height and width of images lie in the range of [700, 3100] and [600, 3500], respectively. 
We cropped the training images into $192\times192$ patches to train all the models. 
Data augmentation was performed by randomly rotating $90^\circ$, $180^\circ$, $270^\circ$ and horizontally flipping the input. The mini-batch size in all the experiments was set to $16$. 

All SISR models were initialized using the method in \cite{he2015delving}. 
The Adam solver \cite{kingma2014adam} with the default parameters ($\beta_1=0.9$, $\beta_2=0.999$ and $\epsilon=10^{-8}$) was adopted to optimize the network parameters. 
The learning rate was fixed at $10^{-4}$ and all the networks were trained for $1,000K$ iterations. 
All the comparing models were trained using the Caffe \cite{jia2014caffe} toolbox, and tested using Caffe MATLAB interface. 
All the experiments were conducted on a PC equipped with an Intel Core i7-7820X CPU, 128G RAM and a single Nvidia Quadro GV100 GPU (32G).
\subsection{Simulated SISR datasets vs. RealSR dataset}
To demonstrate the advantages of our RealSR dataset, we conduct experiments to compare the real-world super-resolution performance of SISR models trained on simulated datasets and RealSR dataset. 
Considering that most state-of-the-art SISR models were trained on DIV2K \cite{timofte2017ntire} dataset, we employed the DIV2K to generate simulated image pairs with bicubic degradation (BD) and multiple degradations (MD) \cite{zhang2018learning}.
We selected three representative and state-of-the-art SISR networks, \ie, VDSR \cite{kim2016accurate}, SRResNet \cite{ledig2017photo} and RCAN \cite{zhang2018image}, and trained them on the BD, MD and RealSR training datasets for each of the three scaling factors ($\times2$, $\times3$, $\times4$), leading to a total of $27$ SISR models. 
To keep the network structures of SRResNet and RCAN unchanged, the input images were shuffled with factor $\frac{1}{2},\frac{1}{3},\frac{1}{4}$ for the three scaling factors $\times2$, $\times3$, $\times4$, respectively. 
\begin{table*}
\footnotesize
\caption{Average PSNR (dB) and SSIM indices on our RealSR testing set by different methods (trained on different datasets).}
\begin{center}
 \begin{tabular}{|c|c|c|c|c|c|c|c|c|c|c|c|}
\hline 
\multirow{2}{*}{Metric}   
&\multirow{2}{*}{Scale}  
&\multirow{2}{*}{Bicubic}  
&\multicolumn{3}{c|}{VDSR \cite{kim2016accurate}}  
&\multicolumn{3}{c|}{SRResNet \cite{ledig2017photo}}  
&\multicolumn{3}{c|}{RCAN \cite{zhang2018image}} \\
\cline{4-12} 
\multicolumn{1}{|c|}{}    
&\multicolumn{1}{c|}{}
&\multicolumn{1}{c|}{}     
&BD  &MD  &\cellcolor{gray!40}Our
&BD  &MD  &\cellcolor{gray!40}Our
&BD  &MD  &\cellcolor{gray!40}Our \\
\hline \hline
\multirow{3}{*}{PSNR}     
&$\times2$     
&$32.61$
&$32.63$   &$32.65$    &\cellcolor{gray!40}$33.64$  
&$32.66$   &$32.69$    &\cellcolor{gray!40}$33.69$ 
&$32.91$   &$32.92$    &\cellcolor{gray!40}$33.87$ \\
\cline{2-2}
&$\times3$
&$29.34$
&$29.40$   &$29.43$    &\cellcolor{gray!40}$30.14$  
&$29.46$   &$29.47$    &\cellcolor{gray!40}$30.18$ 
&$29.66$   &$29.69$    &\cellcolor{gray!40}$30.40$ \\
\cline{2-2}					  
&$\times4$
&$27.99$
&$28.03$   &$28.06$    &\cellcolor{gray!40}$28.63$  
&$28.09$   &$28.12$    &\cellcolor{gray!40}$28.67$ 
&$28.28$   &$28.31$    &\cellcolor{gray!40}$28.88$ \\
\hline \hline
\multirow{3}{*}{SSIM} 
&$\times2$    
&$0.907$
&$0.907$   &$0.908$    &\cellcolor{gray!40}$0.917$  
&$0.908$   &$0.909$    &\cellcolor{gray!40}$0.919$ 
&$0.910$   &$0.912$    &\cellcolor{gray!40}$0.922$ \\
\cline{2-2}
&$\times3$
&$0.841$
&$0.842$   &$0.845$    &\cellcolor{gray!40}$0.856$  
&$0.844$   &$0.846$    &\cellcolor{gray!40}$0.859$ 
&$0.847$   &$0.851$    &\cellcolor{gray!40}$0.862$ \\
\cline{2-2}					  
&$\times4$
&$0.806$
&$0.806$   &$0.807$    &\cellcolor{gray!40}$0.821$  
&$0.806$   &$0.808$    &\cellcolor{gray!40}$0.824$ 
&$0.811$   &$0.813$    &\cellcolor{gray!40}$0.826$ \\
\hline
\end{tabular}
\end{center}
\label{tab:2}
\end{table*}
\begin{figure*}
\scriptsize
\centering
\begin{tabular}{cc}
\begin{adjustbox}{valign=t}
\begin{tabular}{c}
\includegraphics[width=0.23\linewidth]{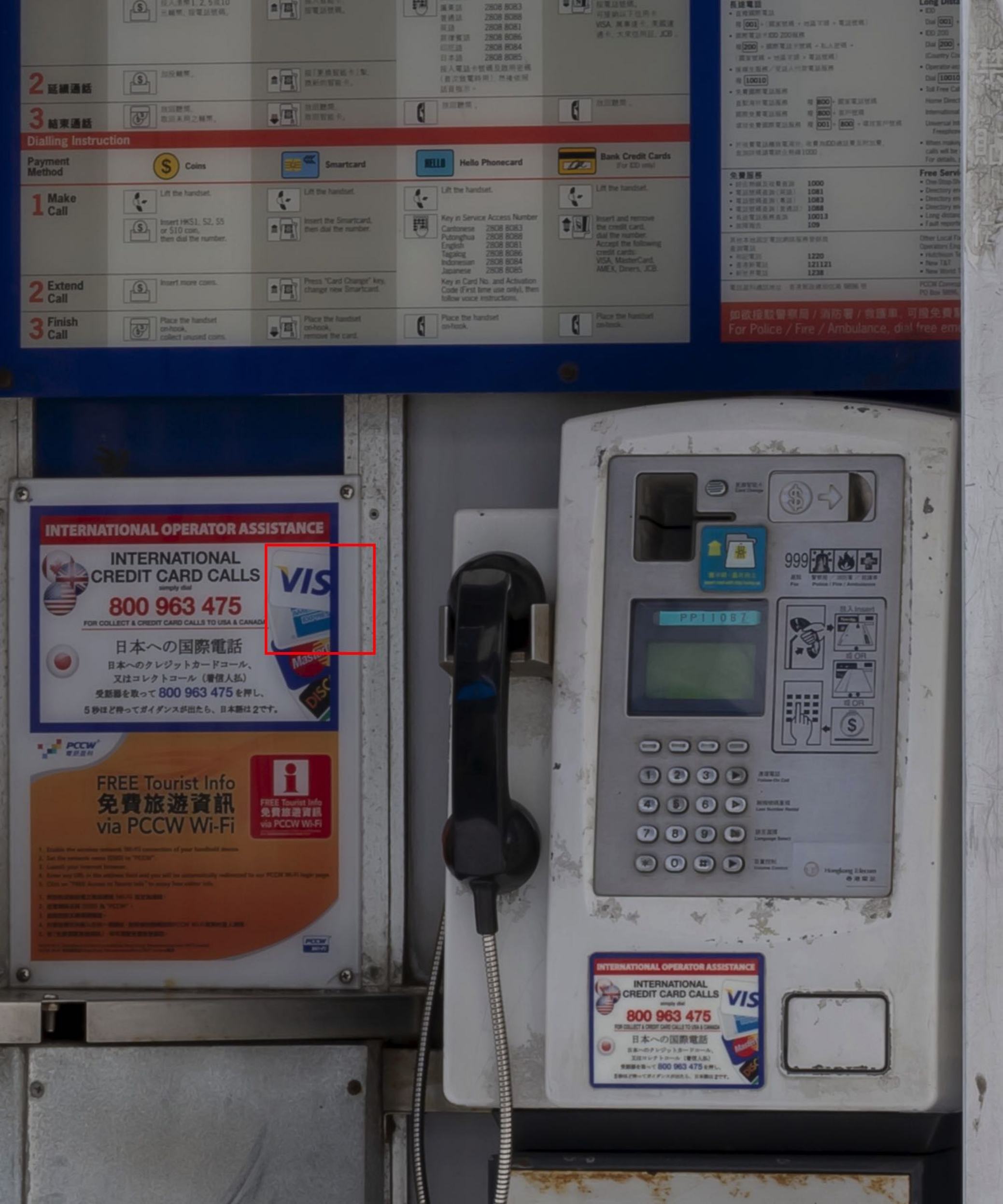}\\
Image captured by Canon 5D3
\end{tabular}
\end{adjustbox}
\scriptsize
\begin{adjustbox}{valign=t}
\begin{tabular}{cccccc}
\includegraphics[width=0.1245\linewidth]{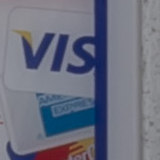}      			&
\includegraphics[width=0.1245\linewidth]{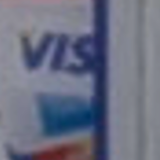} 			&
\includegraphics[width=0.1245\linewidth]{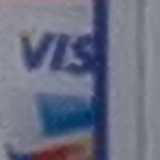}     &
\includegraphics[width=0.1245\linewidth]{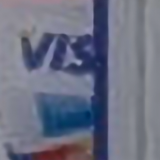}      &
\includegraphics[width=0.1245\linewidth]{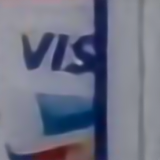}    \\
HR  &Bicubic     &SRResNet + BD    &SRResNet + MD  &SRResNet + RealSR           \\

\includegraphics[width=0.1245\linewidth]{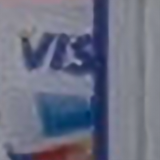}  		&
\includegraphics[width=0.1245\linewidth]{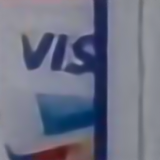}  	&
\includegraphics[width=0.1245\linewidth]{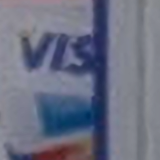}  	&
\includegraphics[width=0.1245\linewidth]{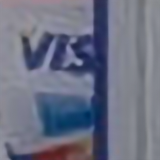}  		&
\includegraphics[width=0.1245\linewidth]{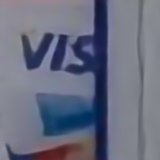}  	\\ 
VDSR + MD    &VDSR + RealSR     &RCAN + BD    &RCAN + MD   &RCAN + RealSR       \\
\end{tabular}
\end{adjustbox}  
\\ \\
\scriptsize
\begin{adjustbox}{valign=t}
\begin{tabular}{c}
\includegraphics[width=0.23\linewidth]{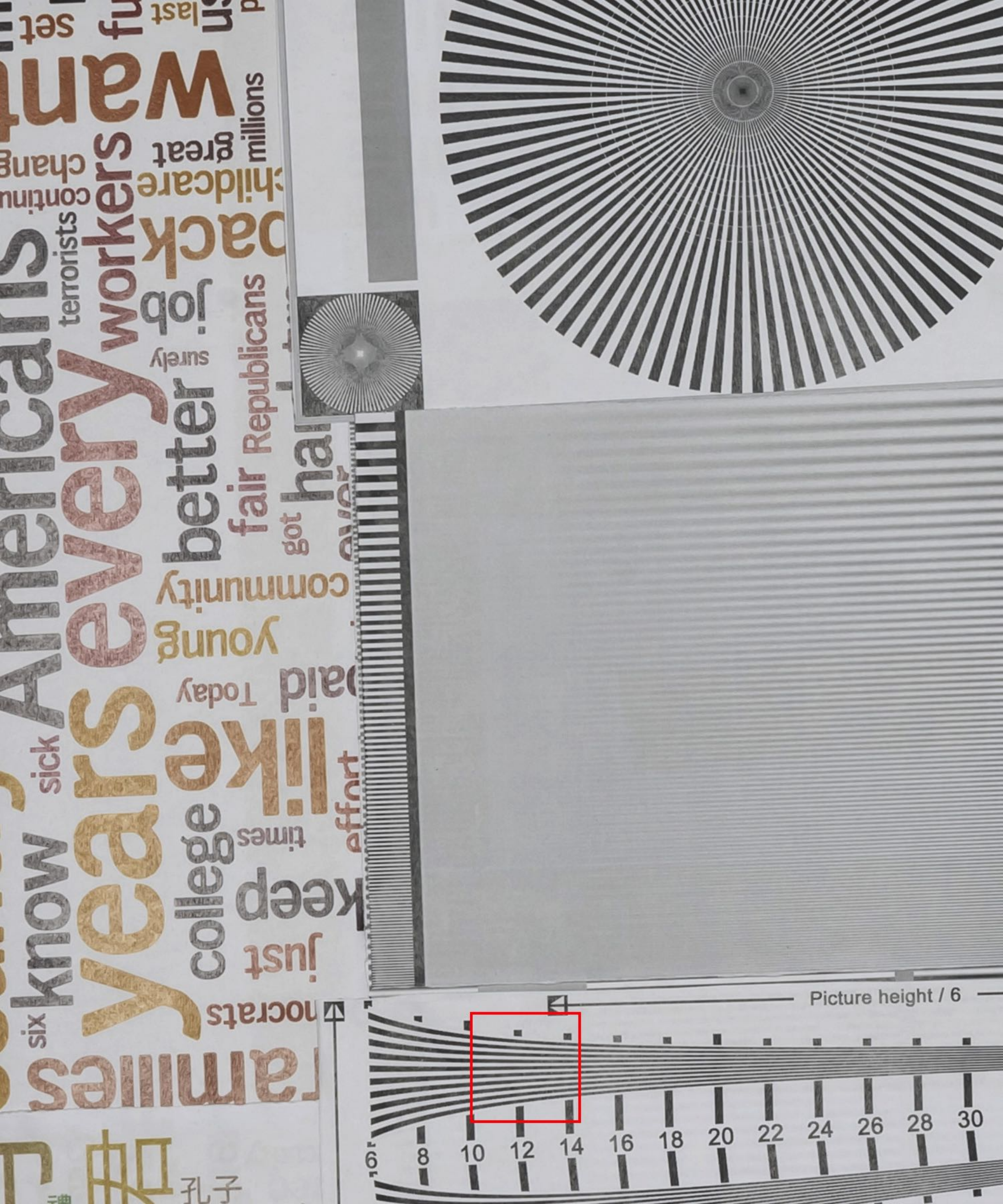}\\
Image captured by Nikon D810
\end{tabular}
\end{adjustbox}
\scriptsize
\begin{adjustbox}{valign=t}
\begin{tabular}{cccccc}
\includegraphics[width=0.1245\linewidth]{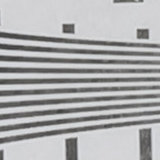}      			&
\includegraphics[width=0.1245\linewidth]{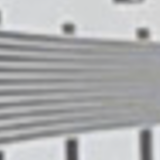} 			&
\includegraphics[width=0.1245\linewidth]{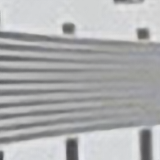}      &
\includegraphics[width=0.1245\linewidth]{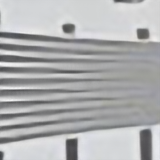}       &
\includegraphics[width=0.1245\linewidth]{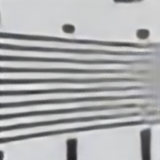}     \\
HR  &Bicubic     &SRResNet + BD    &SRResNet + MD  &SRResNet + RealSR           \\

\includegraphics[width=0.1245\linewidth]{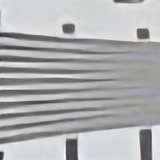}  		&
\includegraphics[width=0.1245\linewidth]{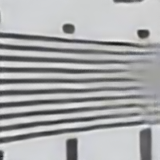}  	&
\includegraphics[width=0.1245\linewidth]{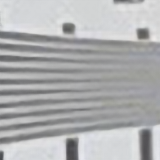}  	    &
\includegraphics[width=0.1245\linewidth]{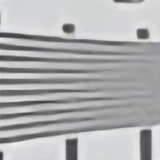}  		&
\includegraphics[width=0.1245\linewidth]{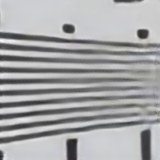}  	\\ 
VDSR + MD    &VDSR + RealSR     &RCAN + BD    &RCAN + MD   &RCAN + RealSR       \\
\end{tabular}
\end{adjustbox} 
\end{tabular}
\caption{SR results ($\times4$) on our RealSR testing set by different methods (trained on different datasets).}
\label{fig:dataset}
\end{figure*}
We applied the $27$ trained SISR models to the RealSR testing set, and the average PSNR and SSIM indices are listed in Table \ref{tab:2}. 
The baseline bicubic interpolator is also included for comparison.
One can see that, on our RealSR testing set, the VDSR and SRResNet models trained on the simulated BD dataset can only achieve comparable performance to the simple bicubic interpolator. 
Training on the MD dataset brings marginal improvements over BD, which indicates that the authentic degradation in real-world images is difficult to simulate. 
Employing a deeper architecture, the RCAN ($>$400 layers) can improve (0.2dB$\sim$0.3dB) the performance over VDSR and SRResNet on all cases. 
Using the same network architecture, SISR models trained on our RealSR dataset obtain significantly better performance than those trained on BD and MD datasets for all the three scaling factors. 
Specifically, for scaling factor $\times 2$, the models trained on our RealSR dataset have about $1.0$dB improvement on average for all the three network architectures. 
The advantage is also significant for scaling factors $\times 3$ and $\times 4$. 
In Fig. \ref{fig:dataset}, we visualize the super-resolved images obtained by different models. 
As can be seen, the SISR results generated by models trained on simulated BD and MD datasets tend to have blurring edges with obvious artifacts. 
On the contrary, models trained on our RealSR dataset recover clearer and more natural image details. More visual examples can be found in the $\textbf{supplementary file}$.
\subsection{SISR models trained on RealSR dataset}
To demonstrate the efficiency and effectiveness of the proposed LP-KPN, we then compare it with $8$ SISR models, including VDSR, SRResNet, RCAN, a baseline direct pixel synthesis (DPS) network and four KPN models with kernel size $k = 5, 7, 13, 19$. 
The DPS and the four KPN models share the same backbone as our LP-KPN. 
All models are trained and tested on our RealSR dataset. 
The PSNR and SSIM indices of all the competing models as well as the bicubic baseline are listed in Table \ref{tab:3}. 
\begin{table}
\footnotesize
\caption{Average PSNR (dB) and SSIM indices for different models (trained on our RealSR training set) on our RealSR testing set.}
\begin{center}
 \begin{tabular}{|l|c|c|c||c|c|c|c|}
\hline 
\multirow{2}{*}{Method}   &\multicolumn{3}{c||}{PSNR} &\multicolumn{3}{c|}{SSIM} \\
\cline{2-7}
\multicolumn{1}{|c|}{}    &$\times2$   &$\times3$   &$\times4$   &$\times2$  &$\times3$   &$\times4$\\
\hline \hline
Bicubic 				  
&$32.61$ &$29.34$ &$27.99$             
&$0.907$ &$0.841$ &$0.806$\\
\hline \hline
VDSR \cite{kim2016accurate}				      
&$33.64$ &$30.14$ &$28.63$             
&$0.917$ &$0.856$ &$0.821$\\
SRResNet \cite{ledig2017photo}			 	  
&$33.69$ &$30.18$ &$28.67$             
&$0.919$ &$0.859$ &$0.824$\\
RCAN \cite{zhang2018image}	  			      
&$33.87$ &$30.40$ &$28.88$             
&$0.922$ &$0.862$ &$0.826$\\
\hline \hline
DPS              
&$33.71$ &$30.20$ &$28.69$                         
&$0.919$ &$0.859$ &$0.824$\\
KPN, $k$ = $5$	              
&$33.75$ &$30.26$ &$28.74$                         
&$0.920$ &$0.860$ &$0.826$\\		
KPN, $k$ = $7$	              
&$33.78$ &$30.29$ &$28.78$                         
&$0.921$ &$0.861$ &$0.827$\\	
KPN, $k$ = $13$	          
&$33.83$ &$30.35$ &$28.85$                         
&$0.923$ &$0.862$ &$0.828$\\			  
KPN, $k$ = $19$		      
&$33.86$ &$30.39$ &$28.90$                         
&$0.924$ &$0.864$ &$0.830$\\
\hline \hline
\rowcolor{gray!40} 
Our, $k$ = $5$                   
&$33.90$ &$30.42$ &$28.92$                         
&$0.927$ &$0.868$ &$0.834$\\
\hline
\end{tabular}
\end{center}
\label{tab:3}
\end{table}
One can notice that among the four direct pixel synthesis networks (\ie, VDSR, SRResNet, RCAN and DPS), RCAN obtains the best performance because of its very deep architecture (over 400 layers). 
Using the same backbone with less than $50$ layers, the KPN with $5\times5$ kernel size already outperforms the DPS.
Using larger kernel size consistently brings better results for the KPN architecture, and it obtains comparable performance to the RCAN when the kernel size increases to $19$. 
Benefitting from the Laplacian pyramid decomposition strategy, our LP-KPN using three different $5\times5$ kernels achieves even better results than the KPN with $19\times19$ kernel. 
The proposed LP-KPN obtains the best performance but with the lowest computational cost for all the three scaling factors. 
The detailed complexity analysis and visual examples of the SISR results by the competing models can be found in the $\textbf{supplementary file}$.
\subsection{Cross-camera testing}
To evaluate the generalization capability of SISR models trained on our RealSR dataset, we conduct a cross-camera testing. 
Images taken by two cameras are divided into training and testing sets, separately, with $15$ testing images for each camera at each scaling factor. 
The three scales of images are combined for training, and models trained on one camera are tested on the testing sets of both cameras. 
The LP-KPN and RCAN models are compared in this evaluation, and the PSNR indexes are reported in Table \ref{tab:4}. 
It can be seen that for both RCAN and LP-KPN, the cross-camera testing results are comparable to the in-camera setting with only about $0.32$dB and $0.30$dB gap, respectively, while both are much better than bicubic interpolator. 
This indicates that the SISR models trained on one camera can generalize well to the other camera. 
This is possibly because our RealSR dataset contains various degradations produced by the camera lens and image formation process, which share similar properties across cameras. 
Between RCAN and LP-KPN models, the former has more parameters and thus is easier to overfit to the training set, delivering slightly worse generalization capability than LP-KPN. 
Similar observation has been found in \cite{bako2017kernel, vogels2018denoising, mildenhall2018burst}.
\begin{table}
\footnotesize
\caption{Average PSNR (dB) index for cross-camera evaluation.}
\begin{center}
 \begin{tabular}{|c|c|c|c|c||c|c|}
\hline 
\multirow{3}{*}{Tested} &\multirow{3}{*}{Scale} &\multirow{3}{*}{Bicubic} &\multicolumn{2}{c||}{RCAN}  &\multicolumn{2}{c|}{LP-KPN} \\
				&   &	 &\multicolumn{2}{c||}{(Trained)}       &\multicolumn{2}{c|}{(Trained)} \\
\cline{4-7}
        		&   &	                            &Canon    &Nikon      &Canon   &Nikon\\
\hline\hline	 
\multirow{3}{*}{Canon}     &$\times2$    &$33.05$   &$34.34$ &$34.11$     &$34.38$ &$34.18$\\

\cline{2-2}				   &$\times3$    &$29.67$   &$30.65$ &$30.28$     &$30.69$ &$30.33$\\

\cline{2-2}				   &$\times4$    &$28.31$   &$29.46$ &$29.04$     &$29.48$ &$29.10$\\
\hline\hline	 
\multirow{3}{*}{Nikon}     &$\times2$    &$31.66$   &$32.01$ &$32.30$     &$32.05$ &$32.33$\\

\cline{2-2}				   &$\times3$    &$28.63$   &$29.30$ &$29.75$     &$29.34$ &$29.78$\\

\cline{2-2}				   &$\times4$    &$27.28$   &$27.98$ &$28.12$     &$28.01$ &$28.13$\\
\hline
\end{tabular}
\end{center}
\label{tab:4}
\end{table}
\begin{figure*}[htp!]
\footnotesize
\centering
\begin{tabular}{cc}
\begin{adjustbox}{valign=t}
\begin{tabular}{c}
\includegraphics[width=0.37\linewidth]{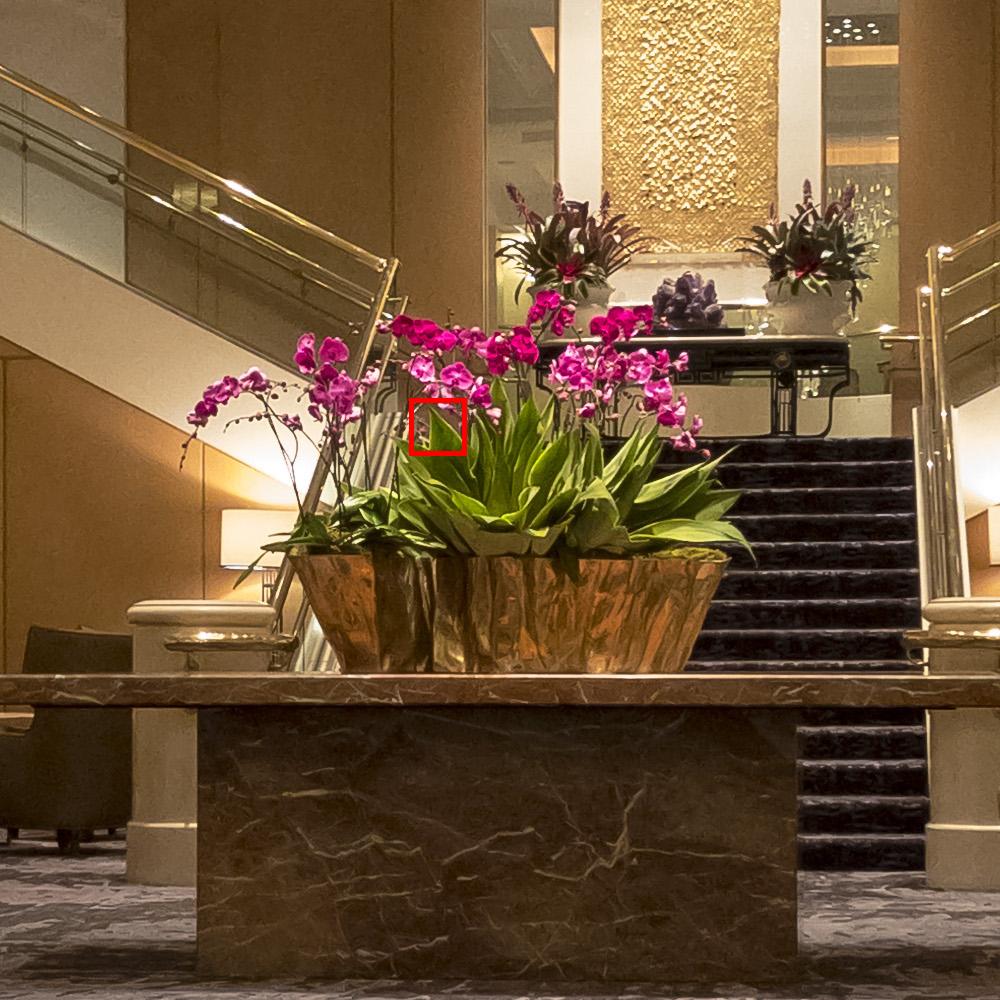}\\
Image captured by iPhone X
\end{tabular}
\end{adjustbox}
\footnotesize
\begin{adjustbox}{valign=t}
\begin{tabular}{ccc}
\includegraphics[width=0.172\linewidth]{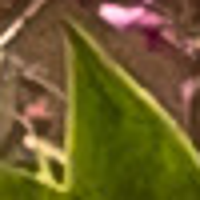}      &
\includegraphics[width=0.172\linewidth]{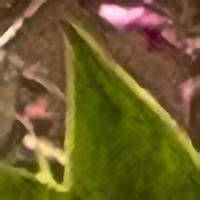} 	      &
\includegraphics[width=0.172\linewidth]{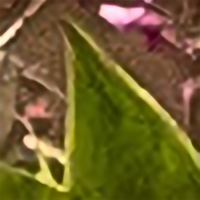}          \\
Bicubic  &RCAN + BD     &RCAN + MD                                        \\

\includegraphics[width=0.172\linewidth]{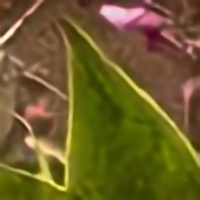}  	       &
\includegraphics[width=0.172\linewidth]{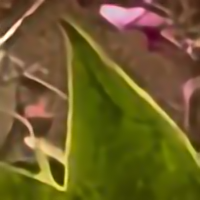}             &
\includegraphics[width=0.172\linewidth]{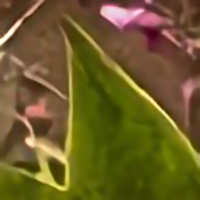}  	      \\ 
RCAN + RealSR    &KPN ($k$ = 19) + RealSR     &LP-KPN + RealSR            \\
\end{tabular}
\end{adjustbox}  
\\ \\
\footnotesize
\begin{adjustbox}{valign=t}
\begin{tabular}{c}
\includegraphics[width=0.37\linewidth]{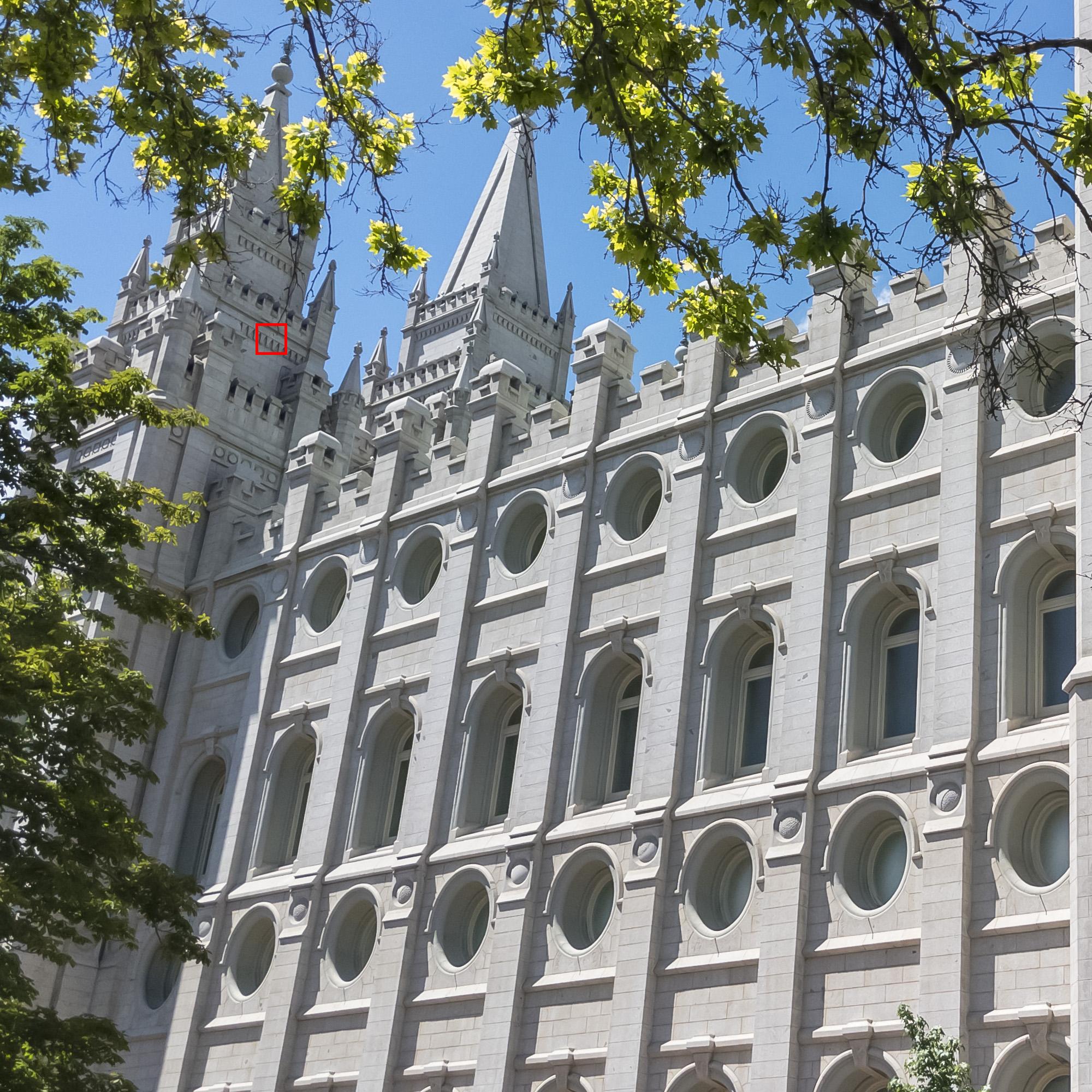}\\
Image captured by Google Pixel 2
\end{tabular}
\end{adjustbox}
\footnotesize
\begin{adjustbox}{valign=t}
\begin{tabular}{ccc}
\includegraphics[width=0.172\linewidth]{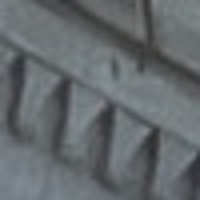}      &
\includegraphics[width=0.172\linewidth]{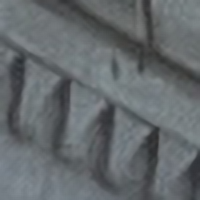} 	      &
\includegraphics[width=0.172\linewidth]{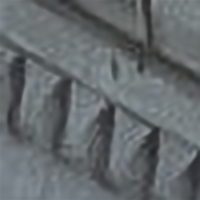}          \\
Bicubic  &RCAN + BD     &RCAN + MD                                        \\

\includegraphics[width=0.172\linewidth]{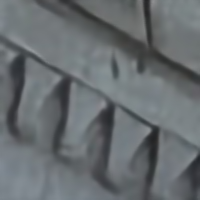}  	      &
\includegraphics[width=0.172\linewidth]{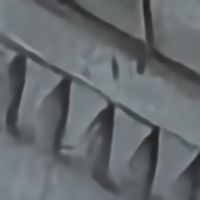}            &
\includegraphics[width=0.172\linewidth]{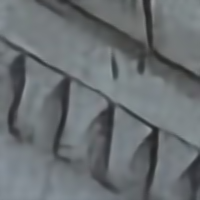}  	      \\ 
RCAN + RealSR    &KPN ($k$ = 19) + RealSR     &LP-KPN + RealSR            \\
\end{tabular}
\end{adjustbox}  
\end{tabular}
\caption{SISR results ($\times4$) of real-world images outside our dataset. Images are captured by iPhone X and Google Pixel 2.}
\label{fig:cross}
\end{figure*}
\subsection{Tests on images outside our dataset}
To further validate the generalization capability of our RealSR dataset and LP-KPN model, we evaluate our trained model as well as several competitors on images outside our dataset, including images taken by one Sony a7II DSLR camera and two mobile cameras (\ie, iPhone X and Google Pixel 2). 
Since there are no ground-truth HR versions of these images, we visualize the super-resolved results in Fig. \ref{figure:SampleIntro} and Fig. \ref{fig:cross}. 
In all these cases, the LP-KPN trained on our RealSR dataset obtains better visual quality than the competitors, recovering more natural and clearer details. 
More examples can be found in the $\textbf{supplementary file}$.
\section{Conclusion}
It has been a long standing problem for SISR research that the models trained on simulated datasets can hardly be generalized to real-world images. 
We made a good attempt to address this issue, and constructed a real-world super-solution (RealSR) dataset with authentic degradations. 
One Canon and one Nikon cameras were used to collect 595 HR and LR image pairs, and an effective image registration algorithm was developed to ensure accurate pixel-wise alignment between image pairs. 
A Laplacian pyramid based kernel prediction network was also proposed to perform efficient and effective real-world SISR. 
Our extensive experiments validated that the models trained on our RealSR dataset can lead to much better real-world SISR results than trained on existing simulated datasets, and they have good generalization capability to other cameras. 
In the future, we will enlarge the RealSR dataset by collecting more image pairs with more types of cameras, and investigate new SISR model training strategies on it.   
\section{Supplementary Material}
\subsection{Sample images of the RealSR dataset}
\begin{figure}[!htp]
\footnotesize
\centering
\subfigure{
\begin{minipage}{0.399\textwidth}
\centering
\includegraphics[width=1\textwidth]{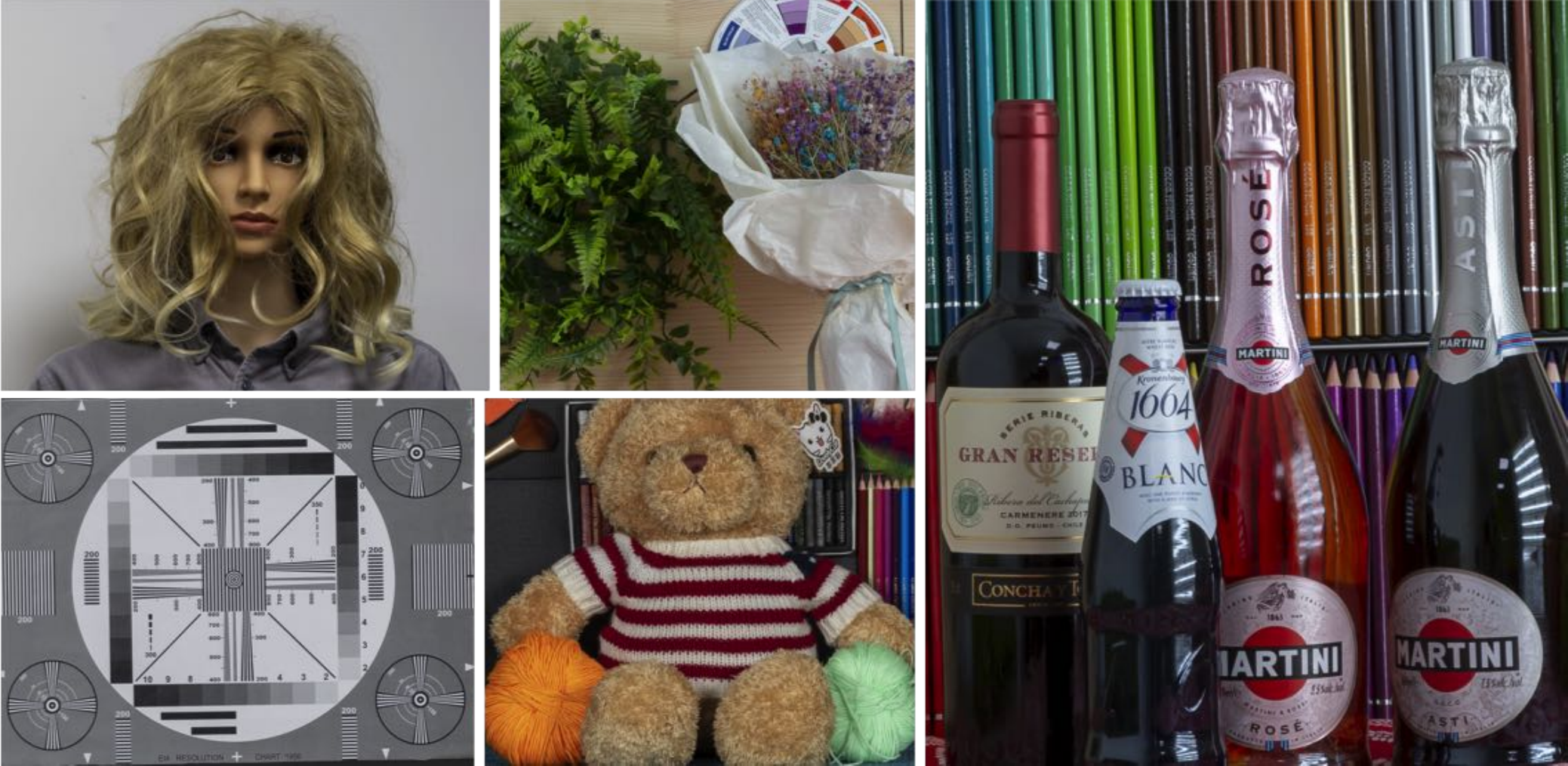}
{(a) Laboratory scenes}
\end{minipage}
}\\
\subfigure{
\begin{minipage}{0.399\textwidth}
\centering
\includegraphics[width=1\textwidth]{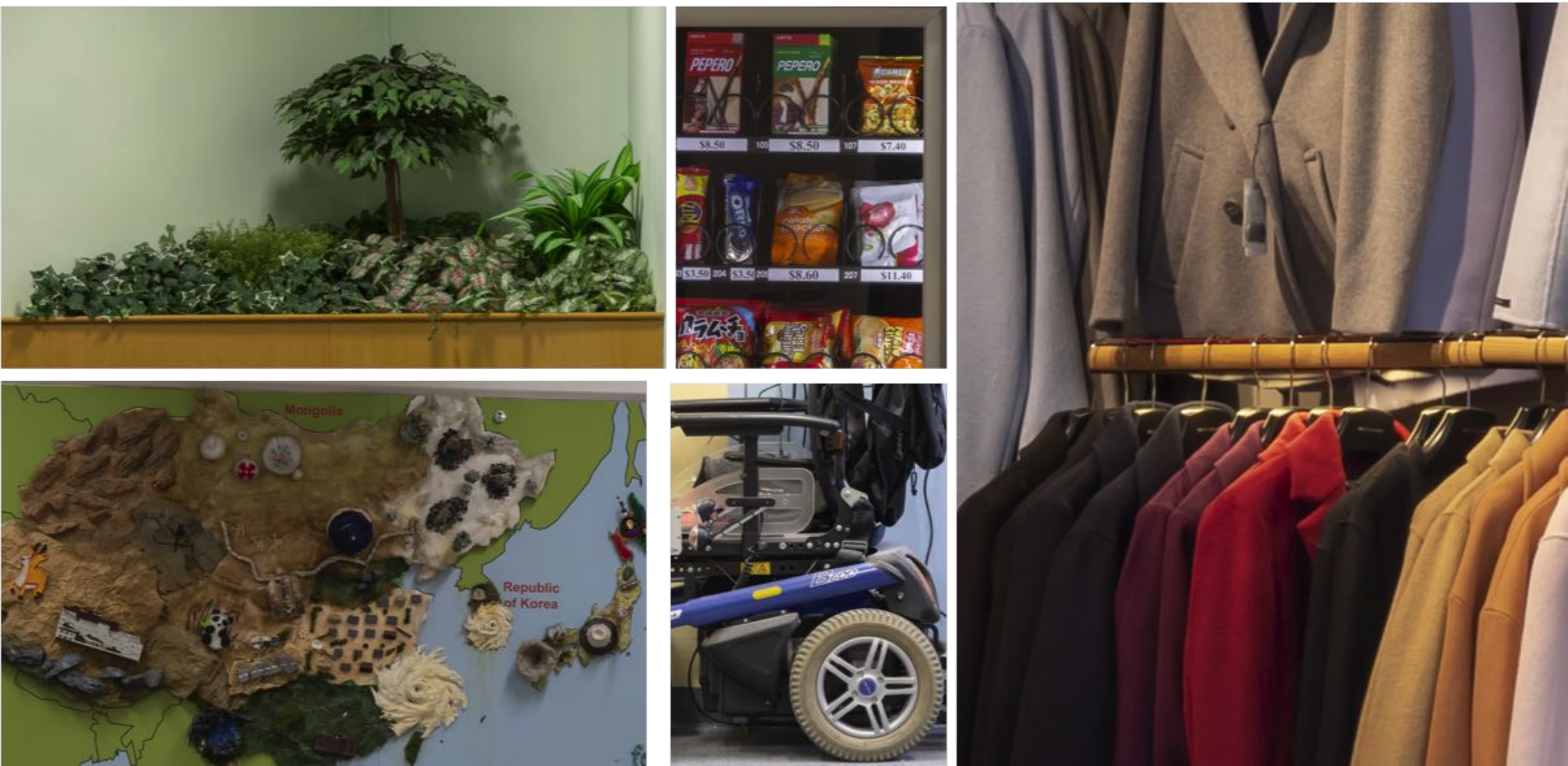}
{(b) Indoor scenes}
\end{minipage}
}\\
\subfigure{
\begin{minipage}{0.399\textwidth}
\centering
\includegraphics[width=1\textwidth]{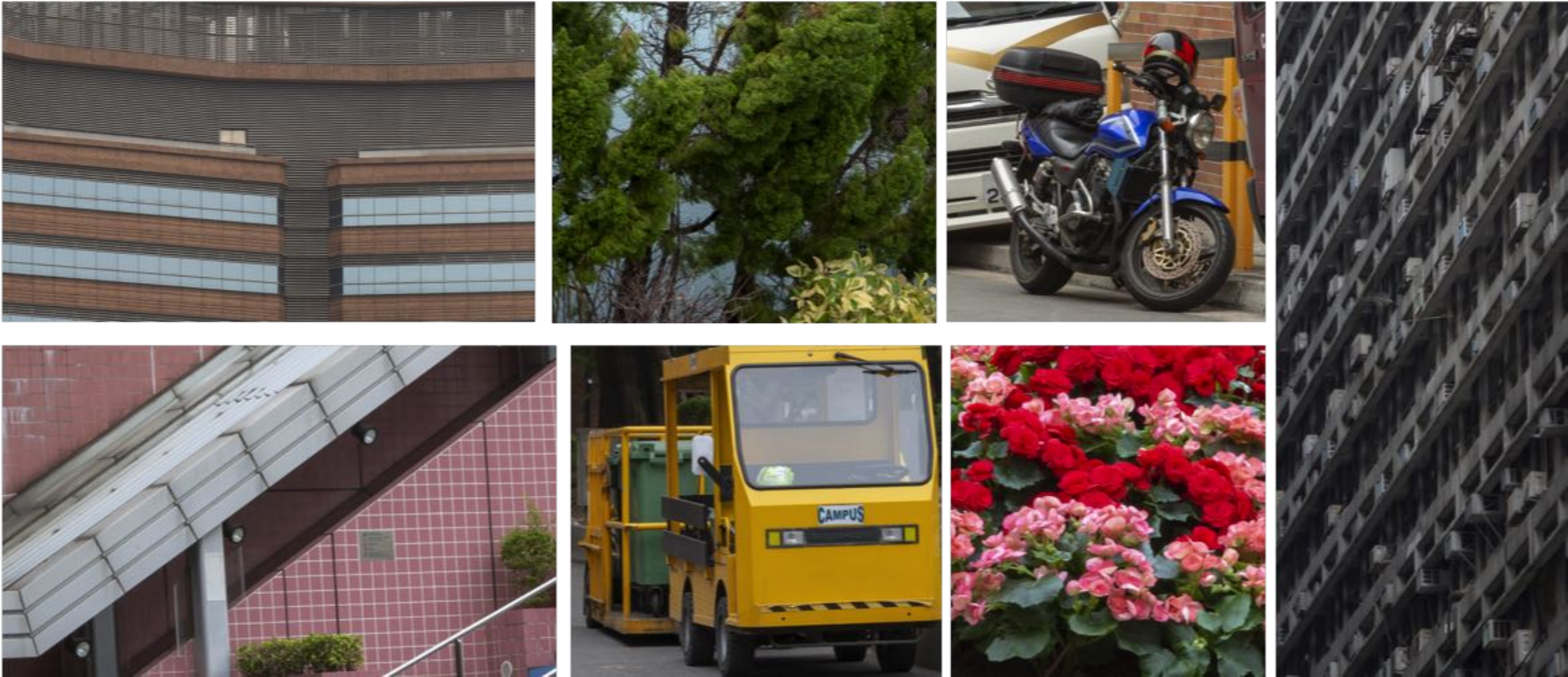}
{(c) Outdoor scenes}
\end{minipage}
}
\caption{Some sample images of our  RealSR dataset.}
\label{sub:sample}
\end{figure}

Currently, the proposed RealSR dataset contains $595$ HR-LR image pairs covering a variety of image contents.
To ensure the diversity of our RealSR dataset, images are captured in indoor, outdoor and laboratory environments. 
Several examples of our RealSR dataset are shown in Fig. \ref{sub:sample}.
It provides, to the best of our knowledge, the first general purpose benchmark for real-world SISR model training and evaluation.
The RealSR dataset will be made publicly available.
\subsection{The details of our network architecture}
\begin{figure}[!htp]
\footnotesize
\centering
\subfigure{
\begin{minipage}{0.55\textwidth}
\centering
\includegraphics[width=1\textwidth]{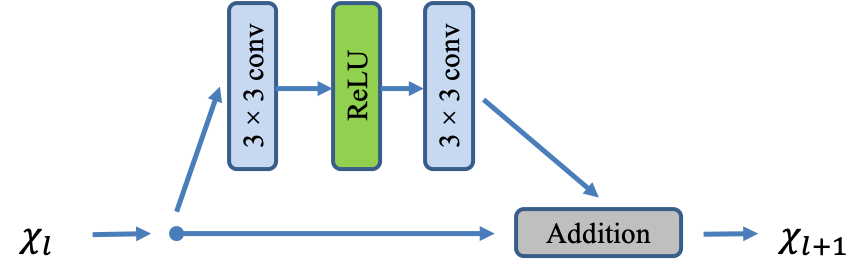}
\end{minipage}
}
\caption{Residual block used in our network.}
\label{sub:net}
\end{figure}

The network architecture of our proposed Laplacian pyramid based kernel prediction network (LP-KPN) is shown in Table \ref{sub:arch}. 
In this table, ``$H \times W \times C$ conv'' denotes a convolutional layer with $C$ filters of size $H \times W$ which is immediately followed by a ReLU nonlinearity.
Each residual block contains two $3\times3$ convolutional layers with the same number of filters on both layers. 
The stride size for all convolution layers is set to $1$ and the number of filters $C$ in each layer is set to $64$, except for the last layer where $C$ is set to $25$.
The structure of the residual block is shown in Fig. \ref{sub:net}, which is same as \cite{lim2017enhanced}.
We use the shuffle operation \cite{shi2016real} to downsample and upsample the the image.
\begin{table*}[htp!]
\centering
\footnotesize
\caption{Network architecture of the proposed LP-KPN.}
\begin{center}
 \begin{tabular}{|c|c|c||c|c|c|}
\hline 
\multicolumn{3}{|c||}{\textbf{Layer}}                          &\multicolumn{3}{c|}{\textbf{Activation size}}\\
\hline \hline
\multicolumn{3}{|c||}{Input}                                    &\multicolumn{3}{c|}{$192\times192$} \\

\multicolumn{3}{|c||}{Shuffle, /4}                              &\multicolumn{3}{c|}{$48\times48\times16$} \\

\multicolumn{3}{|c||}{$3\times3\times64$ conv, pad 1}           &\multicolumn{3}{c|}{$48\times48\times64$} \\

\multicolumn{3}{|c||}{$16\times$ Residual blocks, 64 filters}   &\multicolumn{3}{c|}{$48\times48\times64$} \\

\multicolumn{3}{|c||}{$3\times3\times64$ conv, pad 1}           &\multicolumn{3}{c|}{$48\times48\times64$} \\
\hline
Shuffle, $\times4$  &Shuffle, $\times2$  &-  &$192\times192\times4$  &$96\times96\times16$  &$48\times48\times64$\\
\hline
$3\times3\times64$ conv, pad 1   &$3\times3\times64$ conv, pad 1   &$3\times3\times64$ conv, pad 1 
&$192\times192\times64$          &$96\times96\times64$             &$48\times48\times64$\\

$3\times3\times64$ conv, pad 1   &$3\times3\times64$ conv, pad 1   &$3\times3\times64$ conv, pad 1 
&$192\times192\times64$          &$96\times96\times64$             &$48\times48\times64$\\

$3\times3\times25$ conv, pad 1   &$3\times3\times25$ conv, pad 1   &$3\times3\times25$ conv, pad 1 
&$192\times192\times25$          &$96\times96\times25$             &$48\times48\times25$\\
\hline
Per-pixel conv by Eq. (8)    &Per-pixel conv by Eq. (8)    &Per-pixel conv by Eq. (8) 
&$192\times192$              &$96\times96$  &$48\times48$\\
\hline
\multicolumn{3}{|c||}{Output (Laplacian pyramid reconstruction)}   &\multicolumn{3}{c|}{$192\times192$} \\
\hline
\end{tabular}
\end{center}
\label{sub:arch}
\end{table*}
\begin{figure*}
\scriptsize
\centering
\begin{tabular}{cc}
\begin{adjustbox}{valign=t}
\begin{tabular}{c}
\includegraphics[width=0.21\linewidth]{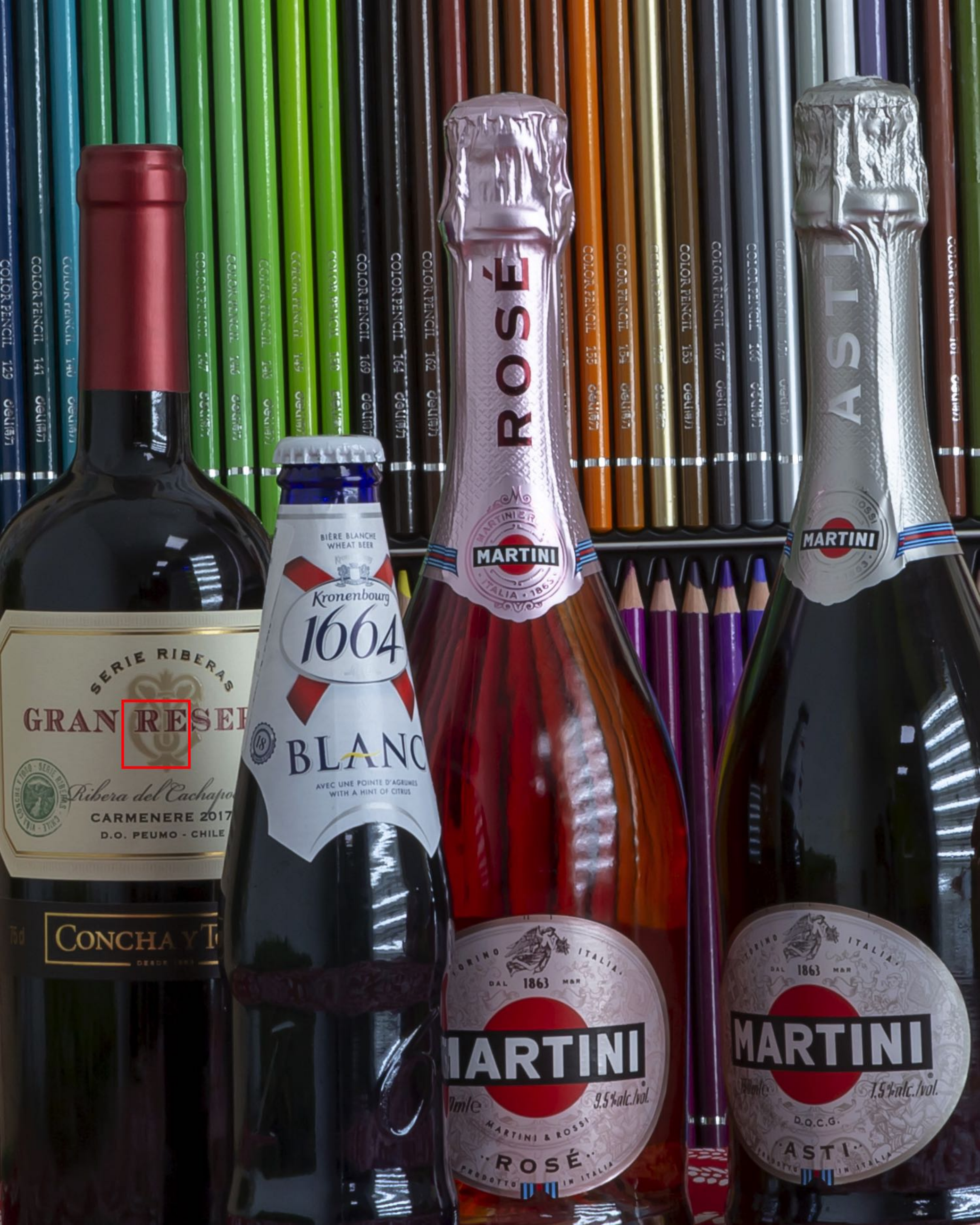}\\
Image captured by Canon 5D3
\end{tabular}
\end{adjustbox}
\scriptsize
\begin{adjustbox}{valign=t}
\begin{tabular}{cccccc}
\includegraphics[width=0.120\linewidth]{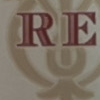}      			&
\includegraphics[width=0.120\linewidth]{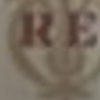} 			&
\includegraphics[width=0.120\linewidth]{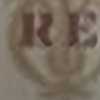}     &
\includegraphics[width=0.120\linewidth]{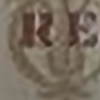}      &
\includegraphics[width=0.120\linewidth]{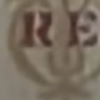}    \\
HR  &Bicubic     &SRResNet + BD    &SRResNet + MD  &SRResNet + RealSR           \\

\includegraphics[width=0.120\linewidth]{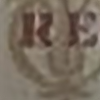}  		&
\includegraphics[width=0.120\linewidth]{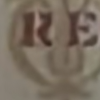}  	&
\includegraphics[width=0.120\linewidth]{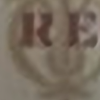}  	     &
\includegraphics[width=0.120\linewidth]{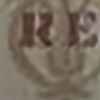}  		&
\includegraphics[width=0.120\linewidth]{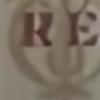}  	\\ 
VDSR + MD    &VDSR + RealSR     &RCAN + BD    &RCAN + MD   &RCAN + RealSR       \\
\end{tabular}
\end{adjustbox}  
\\ \\
\begin{adjustbox}{valign=t}
\begin{tabular}{c}
\includegraphics[width=0.21\linewidth]{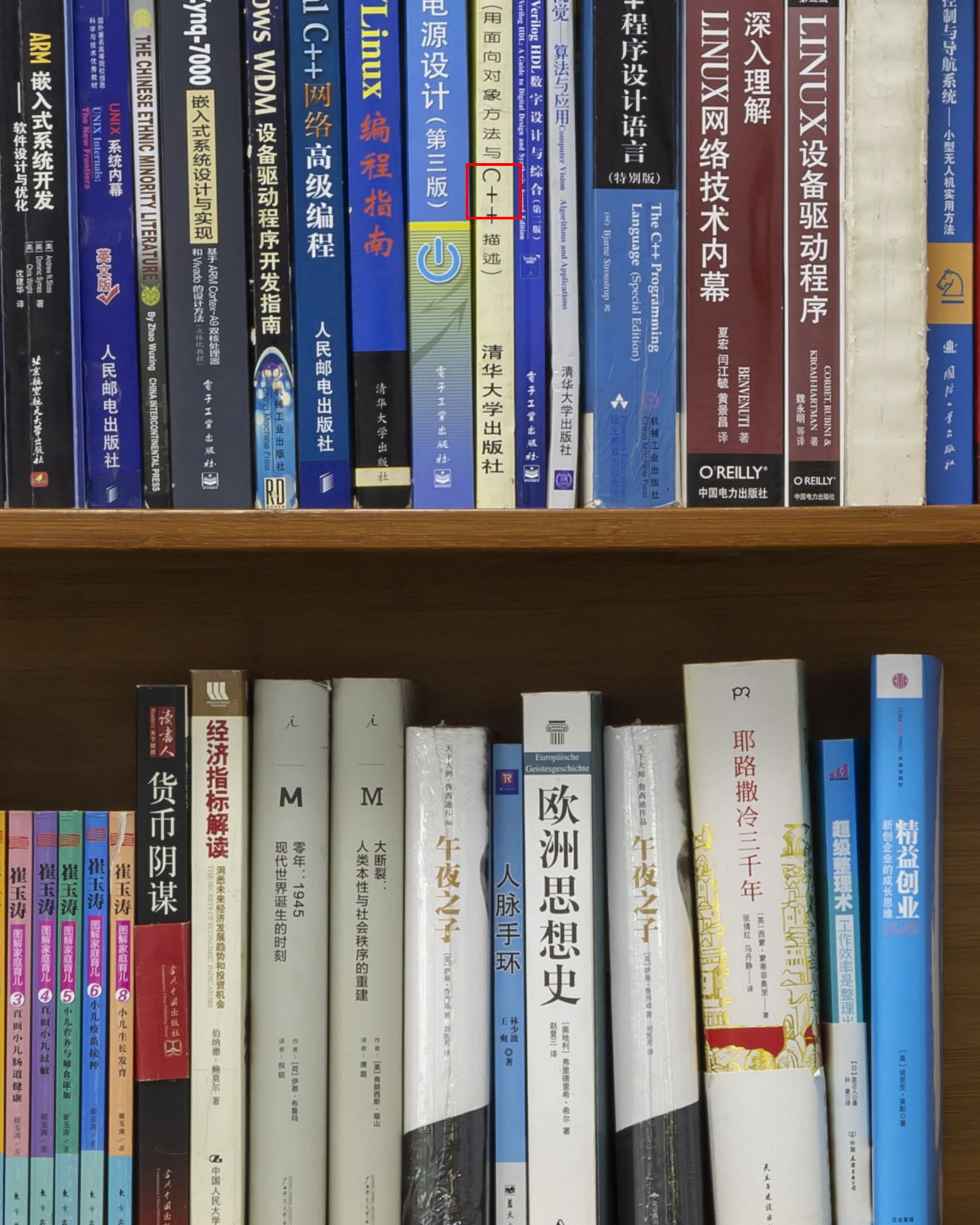}\\
Image captured by Canon 5D3
\end{tabular}
\end{adjustbox}
\scriptsize
\begin{adjustbox}{valign=t}
\begin{tabular}{cccccc}
\includegraphics[width=0.120\linewidth]{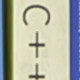}      			&
\includegraphics[width=0.120\linewidth]{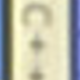} 			&
\includegraphics[width=0.120\linewidth]{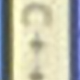}     &
\includegraphics[width=0.120\linewidth]{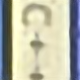}      &
\includegraphics[width=0.120\linewidth]{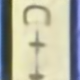}    \\
HR  &Bicubic     &SRResNet + BD    &SRResNet + MD  &SRResNet + RealSR           \\

\includegraphics[width=0.120\linewidth]{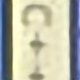}  		&
\includegraphics[width=0.120\linewidth]{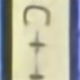}  	&
\includegraphics[width=0.120\linewidth]{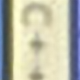}  	     &
\includegraphics[width=0.120\linewidth]{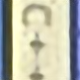}  		&
\includegraphics[width=0.120\linewidth]{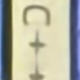}  	\\ 
VDSR + MD    &VDSR + RealSR     &RCAN + BD    &RCAN + MD   &RCAN + RealSR       \\
\end{tabular}
\end{adjustbox}  
\\ \\
\scriptsize
\begin{adjustbox}{valign=t}
\begin{tabular}{c}
\includegraphics[width=0.21\linewidth]{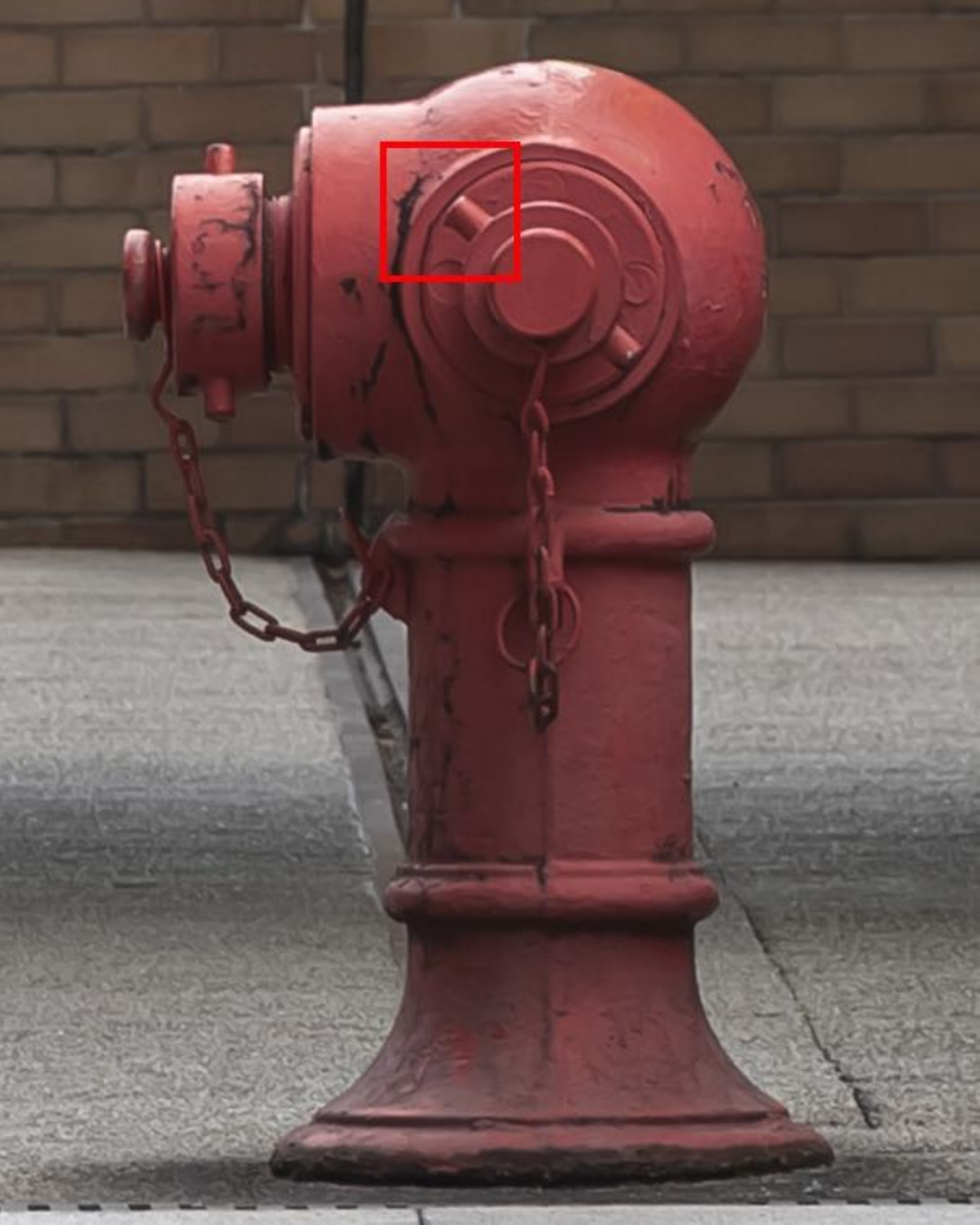}\\
Image captured by Nikon D810
\end{tabular}
\end{adjustbox}
\scriptsize
\begin{adjustbox}{valign=t}
\begin{tabular}{cccccc}
\includegraphics[width=0.120\linewidth]{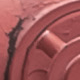}      			&
\includegraphics[width=0.120\linewidth]{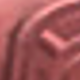} 			&
\includegraphics[width=0.120\linewidth]{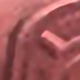}      &
\includegraphics[width=0.120\linewidth]{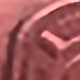}       &
\includegraphics[width=0.120\linewidth]{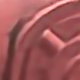}     \\
HR  	&Bicubic            &SRResNet + BD    &SRResNet + MD   &SRResNet + RealSR  \\

\includegraphics[width=0.120\linewidth]{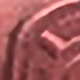}  		&
\includegraphics[width=0.120\linewidth]{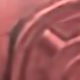}  	&
\includegraphics[width=0.120\linewidth]{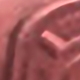}  	    &
\includegraphics[width=0.120\linewidth]{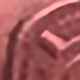}  		&
\includegraphics[width=0.120\linewidth]{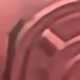}  	\\ 
VDSR + MD    &VDSR + RealSR     &RCAN + BD        &RCAN + MD        &RCAN + RealSR      \\
\end{tabular}
\end{adjustbox} 
\\ \\
\scriptsize
\begin{adjustbox}{valign=t}
\begin{tabular}{c}
\includegraphics[width=0.21\linewidth]{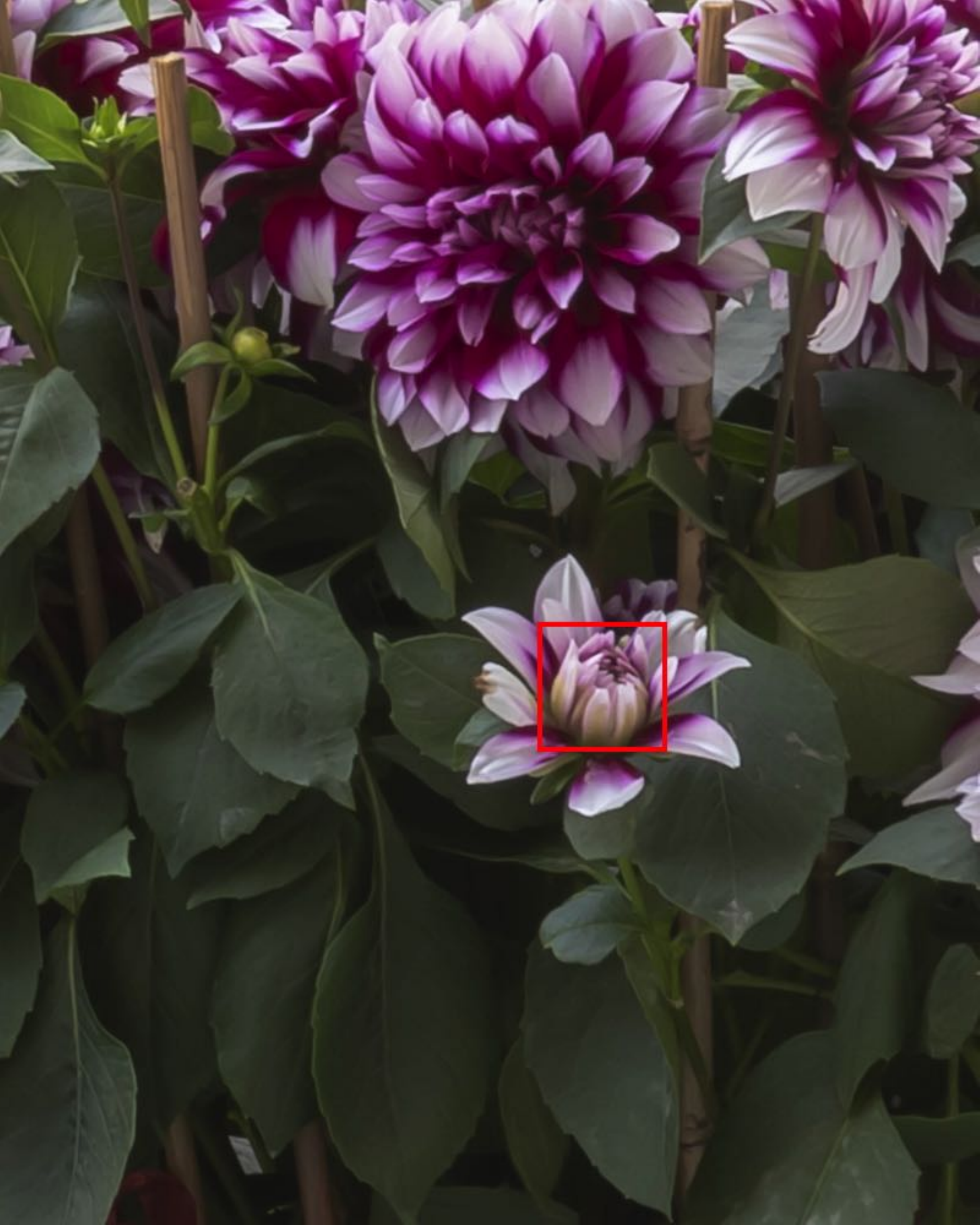}\\
Image captured by Nikon D810
\end{tabular}
\end{adjustbox}
\scriptsize
\begin{adjustbox}{valign=t}
\begin{tabular}{cccccc}
\includegraphics[width=0.120\linewidth]{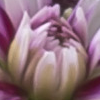}      			&
\includegraphics[width=0.120\linewidth]{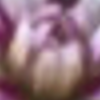} 			&
\includegraphics[width=0.120\linewidth]{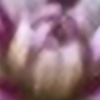}      &
\includegraphics[width=0.120\linewidth]{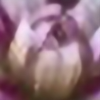}       &
\includegraphics[width=0.120\linewidth]{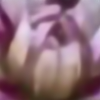}     \\
HR  	&Bicubic            &SRResNet + BD    &SRResNet + MD   &SRResNet + RealSR  \\

\includegraphics[width=0.120\linewidth]{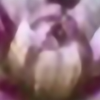}  		&
\includegraphics[width=0.120\linewidth]{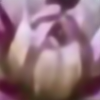}  	&
\includegraphics[width=0.120\linewidth]{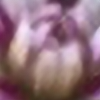}  	    &
\includegraphics[width=0.120\linewidth]{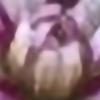}  		&
\includegraphics[width=0.120\linewidth]{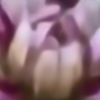}  	\\ 
VDSR + MD    &VDSR + RealSR     &RCAN + BD        &RCAN + MD        &RCAN + RealSR      \\
\end{tabular}
\end{adjustbox} 
\end{tabular}
\caption{SR results ($\times4$) on our RealSR testing set by different methods (trained on different datasets).}
\label{sub:dataset}
\end{figure*}
\subsection{More visual results by SISR models trained on simulated SISR datasets and our RealSR dataset}
In this subsection, we provide more visual results by SISR models trained on simulated SISR datasets (BD and MD \cite{zhang2018learning}) and our proposed RealSR dataset. Two images captured by Canon 5D3, two images captured by Nikon D810 and their super-resolved results are shown in Fig. \ref{sub:dataset}.
Again, the models trained on our RealSR dataset consistently obtain better visual quality compared to their counterparts trained on simulated datasets.
\subsection{The computational cost of competing models and their visual results when trained on our RealSR dataset}
\begin{figure*}[htp!]
\footnotesize
\centering
\begin{tabular}{cc}
\begin{adjustbox}{valign=t}
\begin{tabular}{c}
\includegraphics[width=0.29\linewidth]{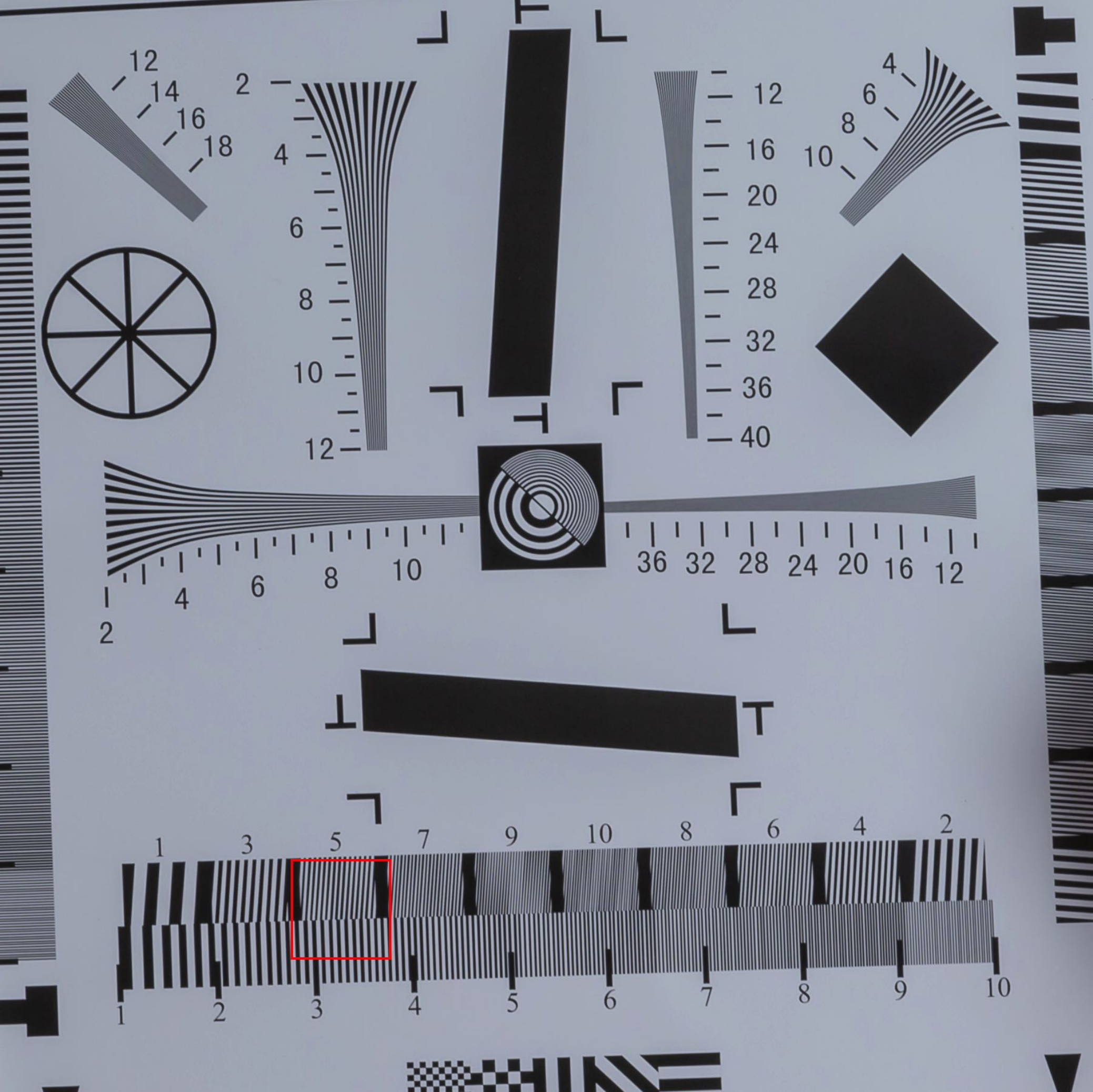}\\
Image captured by Canon 5D3
\end{tabular}
\end{adjustbox}
\footnotesize
\begin{adjustbox}{valign=t}
\begin{tabular}{ccccc}
\includegraphics[width=0.133\linewidth]{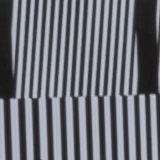}       &
\includegraphics[width=0.133\linewidth]{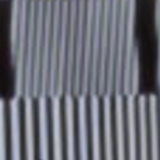}  &
\includegraphics[width=0.133\linewidth]{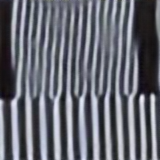}     &
\includegraphics[width=0.133\linewidth]{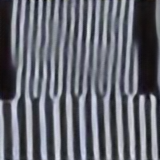}    \\
HR        &Bicubic             &VDSR                &SRResNet         \\

\includegraphics[width=0.133\linewidth]{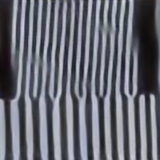}     &
\includegraphics[width=0.133\linewidth]{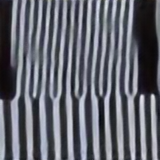}   &
\includegraphics[width=0.133\linewidth]{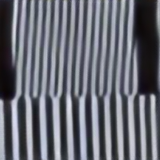}      &
\includegraphics[width=0.133\linewidth]{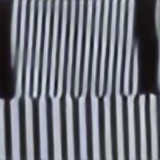}      \\ 
RCAN       &Direct Synth       &KPN ($k=19$)        &Our              \\
\end{tabular}
\end{adjustbox}
\\ 
% \\ 
% \begin{adjustbox}{valign=t}
% \begin{tabular}{c}
% \includegraphics[width=0.29\linewidth]{Sup/network3/Canon_005_HR.png}\\
% Image captured by Canon 5D3
% \end{tabular}
% \end{adjustbox}
% \footnotesize
% \begin{adjustbox}{valign=t}
% \begin{tabular}{ccccc}
% \includegraphics[width=0.133\linewidth]{Sup/network3/HR.png}       &
% \includegraphics[width=0.133\linewidth]{Sup/network3/bicubic.png}  &
% \includegraphics[width=0.133\linewidth]{Sup/network3/VDSR.png}     &
% \includegraphics[width=0.133\linewidth]{Sup/network3/SRRes.png}    \\
% HR        &Bicubic             &VDSR                &SRResNet         \\

% \includegraphics[width=0.133\linewidth]{Sup/network3/RCAN.png}     &
% \includegraphics[width=0.133\linewidth]{Sup/network3/Direct.png}   &
% \includegraphics[width=0.133\linewidth]{Sup/network3/KPN.png}      &
% \includegraphics[width=0.133\linewidth]{Sup/network3/Our.png}      \\ 
% RCAN       &Direct Synth       &KPN ($k=19$)        &Our              \\
% \end{tabular}
% \end{adjustbox}
% \\ 
\\ 
\begin{adjustbox}{valign=t}
\begin{tabular}{c}
\includegraphics[width=0.29\linewidth]{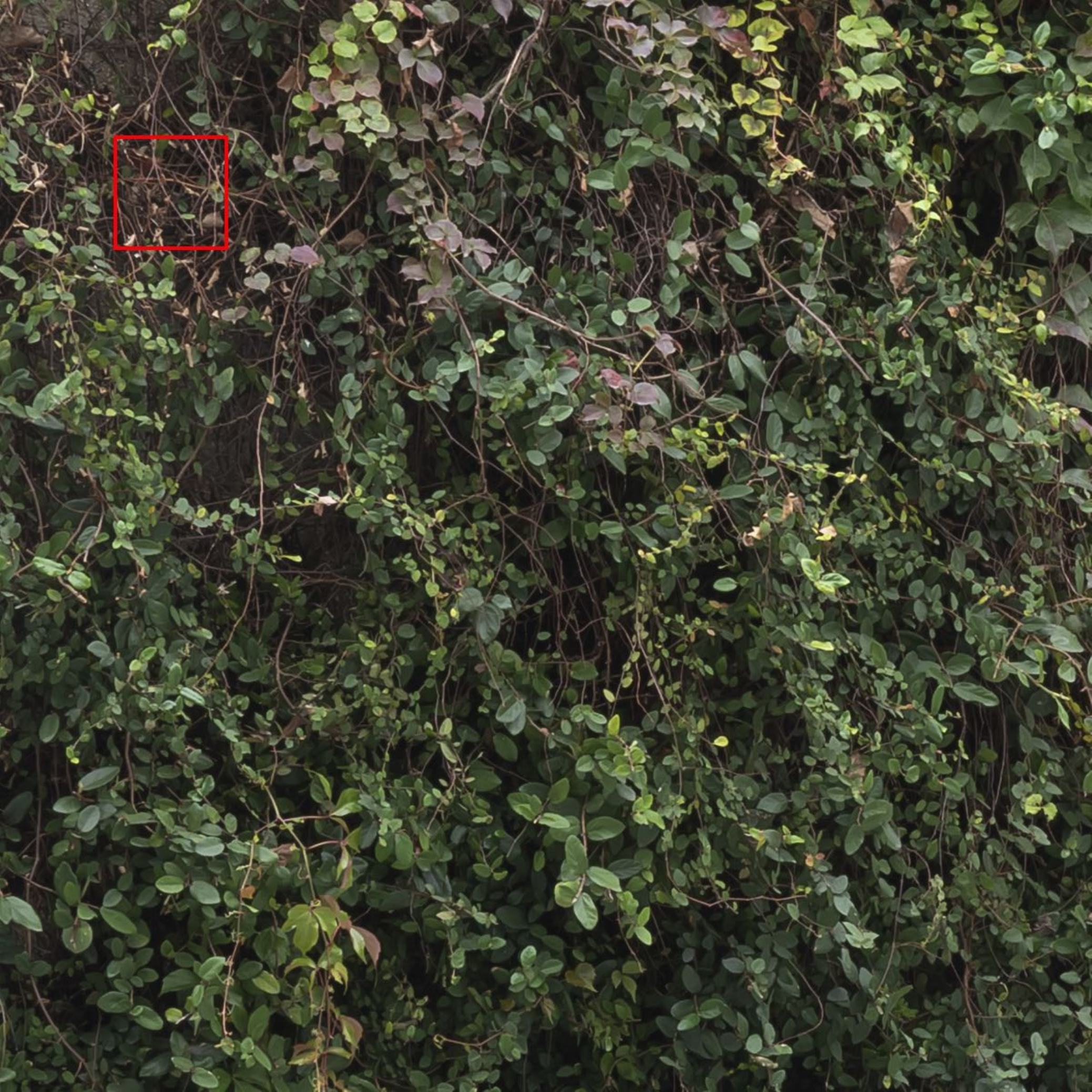}\\
Image captured by Nikon D810
\end{tabular}
\end{adjustbox}
\footnotesize
\begin{adjustbox}{valign=t}
\begin{tabular}{ccccc}
\includegraphics[width=0.133\linewidth]{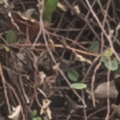}       &
\includegraphics[width=0.133\linewidth]{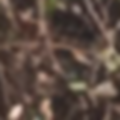}  &
\includegraphics[width=0.133\linewidth]{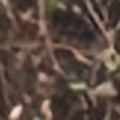}     &
\includegraphics[width=0.133\linewidth]{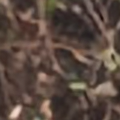}    \\
HR        &Bicubic             &VDSR                &SRResNet         \\

\includegraphics[width=0.133\linewidth]{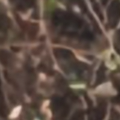}     &
\includegraphics[width=0.133\linewidth]{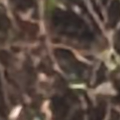}   &
\includegraphics[width=0.133\linewidth]{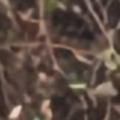}      &
\includegraphics[width=0.133\linewidth]{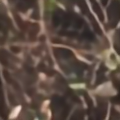}      \\ 
RCAN       &Direct Synth       &KPN ($k=19$)        &Our              \\
\end{tabular}
\end{adjustbox}
\end{tabular}
\caption{SR results ($\times3$) on our RealSR testing set by different methods (all trained on our RealSR dataset).
It can be seen that all SISR models trained on our RealSR dataset achieve good results, while our LP-KPN still obtains the best visual quality.}
\label{sub:network}
\end{figure*}

The running time and the number of parameters of the competing models are listed in Table \ref{sub:par}.
One can see that although larger kernel size can consistently bring better results for the KPN architecture, the number of parameters will also greatly increase.
Benefitting from the Laplacian pyramid decomposition strategy, our LP-KPN using $5\times5$ kernel can achieve better results than the KPN using $19\times19$ kernel, and it uses much less parameters. 
Specifically, our LP-KPN model contains less than $\frac{1}{5}$ parameters of the RCAN model \cite{zhang2018image} and it runs about 3 times faster than RCAN.
The visual examples of the SISR results by the competing models are shown in Fig. \ref{sub:network}.
Though all the SISR models in Fig. \ref{sub:network} are trained on our RealSR dataset and they all achieve good results, our LP-KPN still obtains the best visual quality among the competitors.
\begin{table*}[!htp]
\scriptsize
\caption{PSNR, SSIM, running time and parameters for different models (trained on our RealSR training set) on our RealSR testing set. 
The running time is measured for an image of size $1200\times2200$.
We use the file size of Caffe model to represent the number of parameters.}
\begin{center}
 \begin{tabular}{|l|c||c|c|c||c|c|c|c|c||c|}
\hline 
\multicolumn{2}{|c||}{ }  &VDSR \cite{kim2016accurate} &SRResNet \cite{ledig2017photo}	 &RCAN \cite{zhang2018image}	&DPS &KPN, $k$ = $5$	 &KPN, $k$ = $7$	 &KPN, $k$ = $13$	 &KPN, $k$ = $19$	 &Our, $k$ = $5$	\\
\hline
\multirow{3}{*}{PSNR}  &$\times2$    &$33.64$  &$33.69$  &$33.87$  &$33.71$ &$33.75$ &$33.78$ &$33.83$ &$33.86$ &$33.90$\\
\cline{2-2}			   &$\times3$    &$30.14$  &$30.18$  &$30.40$  &$30.20$ &$30.26$ &$30.29$ &$30.35$ &$30.39$ &$30.42$\\
\cline{2-2} 		   &$\times4$    &$28.63$  &$28.67$  &$28.88$  &$28.69$ &$28.74$ &$28.78$ &$28.85$ &$28.90$ &$28.92$\\
\hline \hline
\multirow{3}{*}{SSIM}  &$\times2$    &$0.917$  &$0.919$  &$0.922$  &$0.919$ &$0.920$ &$0.921$ &$0.923$ &$0.924$ &$0.927$\\
\cline{2-2}			   &$\times3$    &$0.856$  &$0.859$  &$0.862$  &$0.859$ &$0.860$ &$0.861$ &$0.862$ &$0.864$ &$0.868$\\
\cline{2-2} 		   &$\times4$    &$0.821$  &$0.824$  &$0.826$  &$0.824$ &$0.826$ &$0.827$ &$0.828$ &$0.830$ &$0.834$\\
\hline \hline
\multicolumn{2}{|c||}{Parameters}    &$2.667$M &$5.225$M &$32.71$M &$5.079$M &$5.134$M &$5.190$M  &$5.467$M &$5.910$M &$5.731$M\\
\hline
\multicolumn{2}{|c||}{Times (sec.)}  &$0.4262$ &$0.1431$ &$0.5106$ &$0.1228$ &$0.1296$ &$0.1391$ &$0.1909$ &$0.2748$ &$0.1813$\\
\hline
\end{tabular}
\end{center}
\label{sub:par}
\end{table*}
\begin{figure*}[!]
\scriptsize
\centering
\begin{tabular}{cc}
\scriptsize
\begin{adjustbox}{valign=t}
\begin{tabular}{c}
\includegraphics[width=0.37\linewidth]{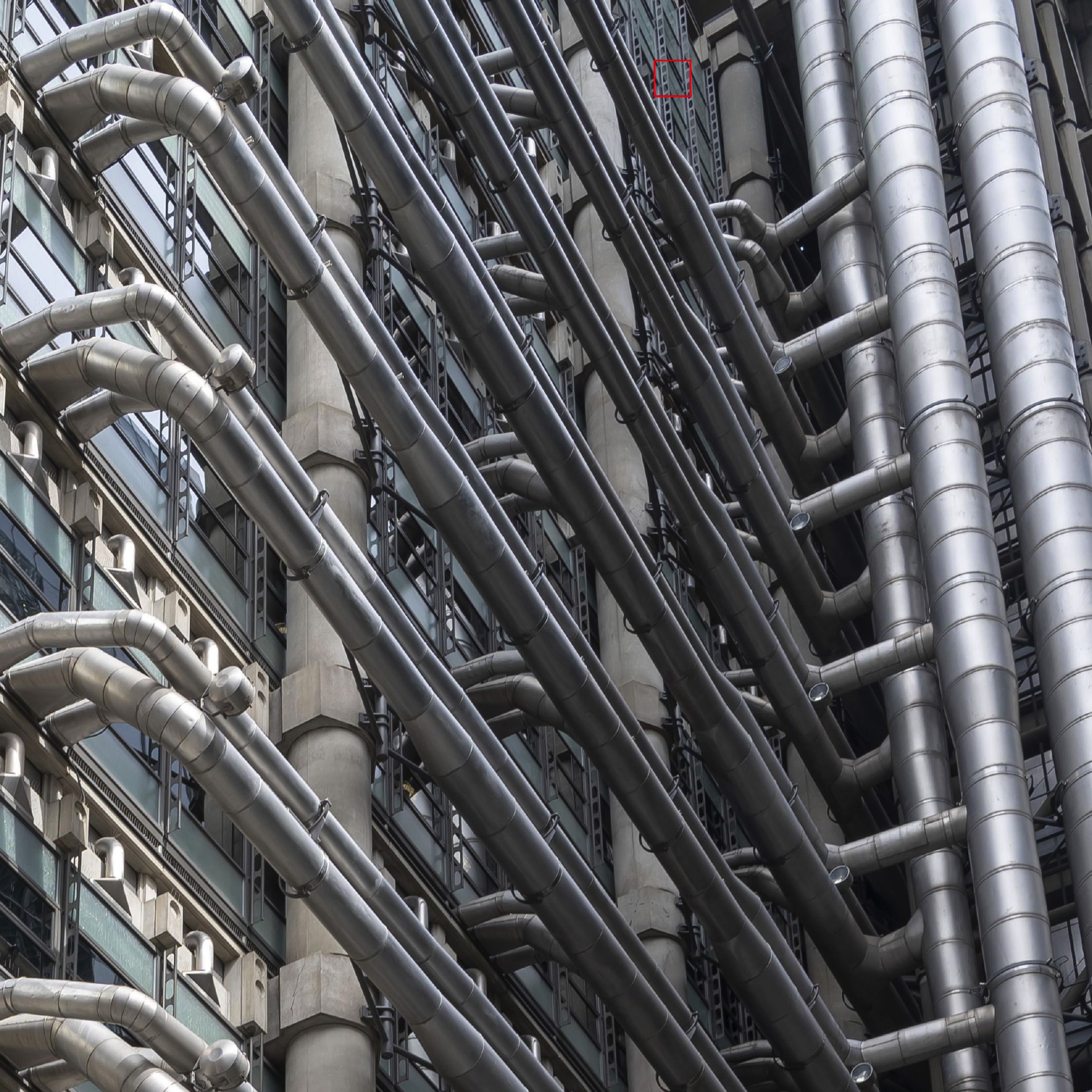}\\
Image captured by Sony a7II
\end{tabular}
\end{adjustbox}
\scriptsize
\begin{adjustbox}{valign=t}
\begin{tabular}{ccc}
\includegraphics[width=0.172\linewidth]{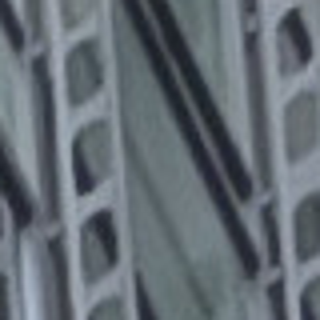}      &
\includegraphics[width=0.172\linewidth]{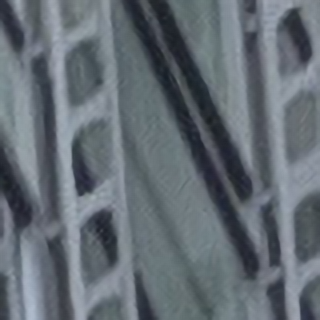} 	      &
\includegraphics[width=0.172\linewidth]{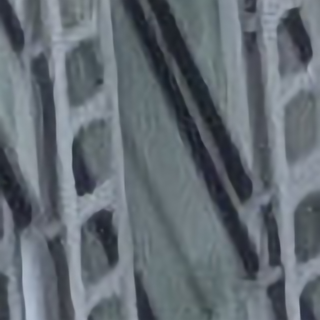}          \\
Bicubic  &RCAN + BD     &RCAN + MD                                        \\

\includegraphics[width=0.172\linewidth]{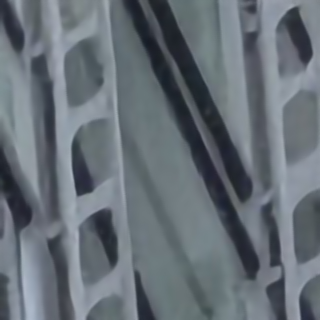}  	      &
\includegraphics[width=0.172\linewidth]{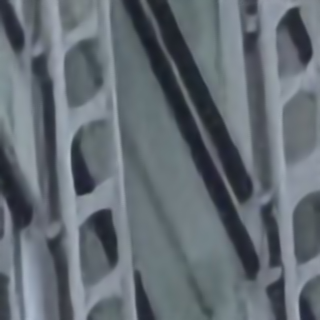}            &
\includegraphics[width=0.172\linewidth]{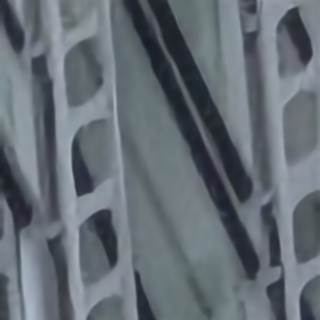}  	      \\ 
RCAN + RealSR    &KPN ($k$ = 19) + RealSR     &LP-KPN + RealSR            \\
\end{tabular}
\end{adjustbox}
\\ 
\scriptsize
\begin{adjustbox}{valign=t}
\begin{tabular}{c}
\includegraphics[width=0.37\linewidth]{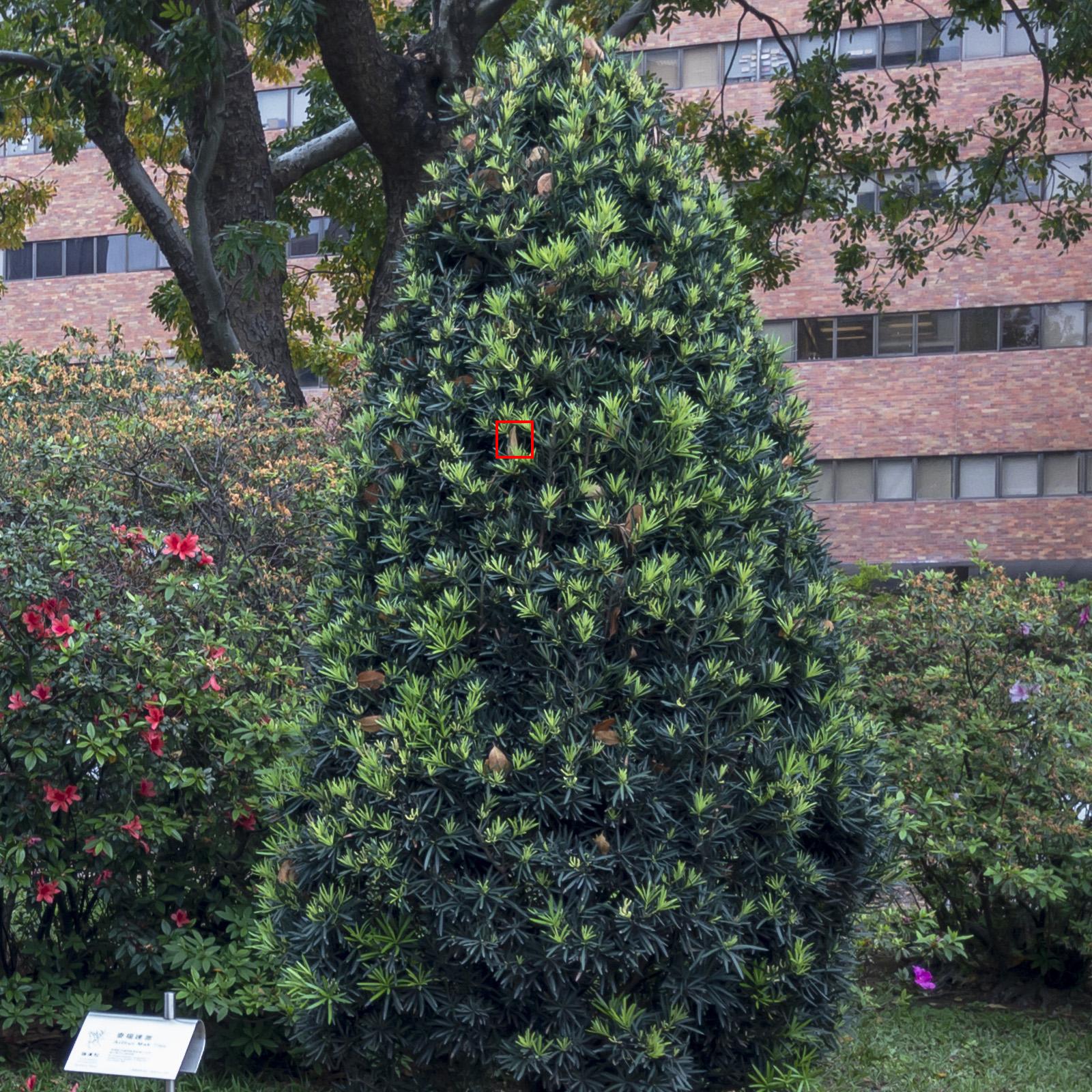}\\
Image captured by iPhone X
\end{tabular}
\end{adjustbox}
\scriptsize
\begin{adjustbox}{valign=t}
\begin{tabular}{ccc}
\includegraphics[width=0.172\linewidth]{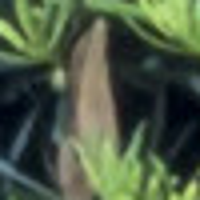}      &
\includegraphics[width=0.172\linewidth]{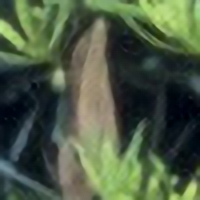} 	      &
\includegraphics[width=0.172\linewidth]{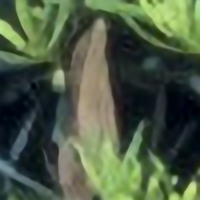}          \\
Bicubic  &RCAN + BD     &RCAN + MD                                        \\

\includegraphics[width=0.172\linewidth]{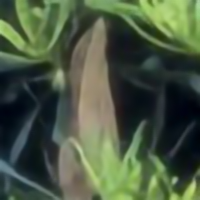}  	       &
\includegraphics[width=0.172\linewidth]{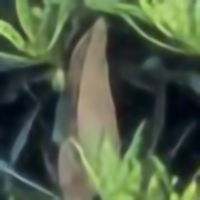}             &
\includegraphics[width=0.172\linewidth]{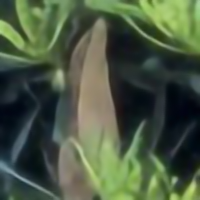}  	      \\ 
RCAN + RealSR    &KPN ($k$ = 19) + RealSR     &LP-KPN + RealSR            \\
\end{tabular}
\end{adjustbox}  
\\ \\
\scriptsize
\begin{adjustbox}{valign=t}
\begin{tabular}{c}
\includegraphics[width=0.37\linewidth]{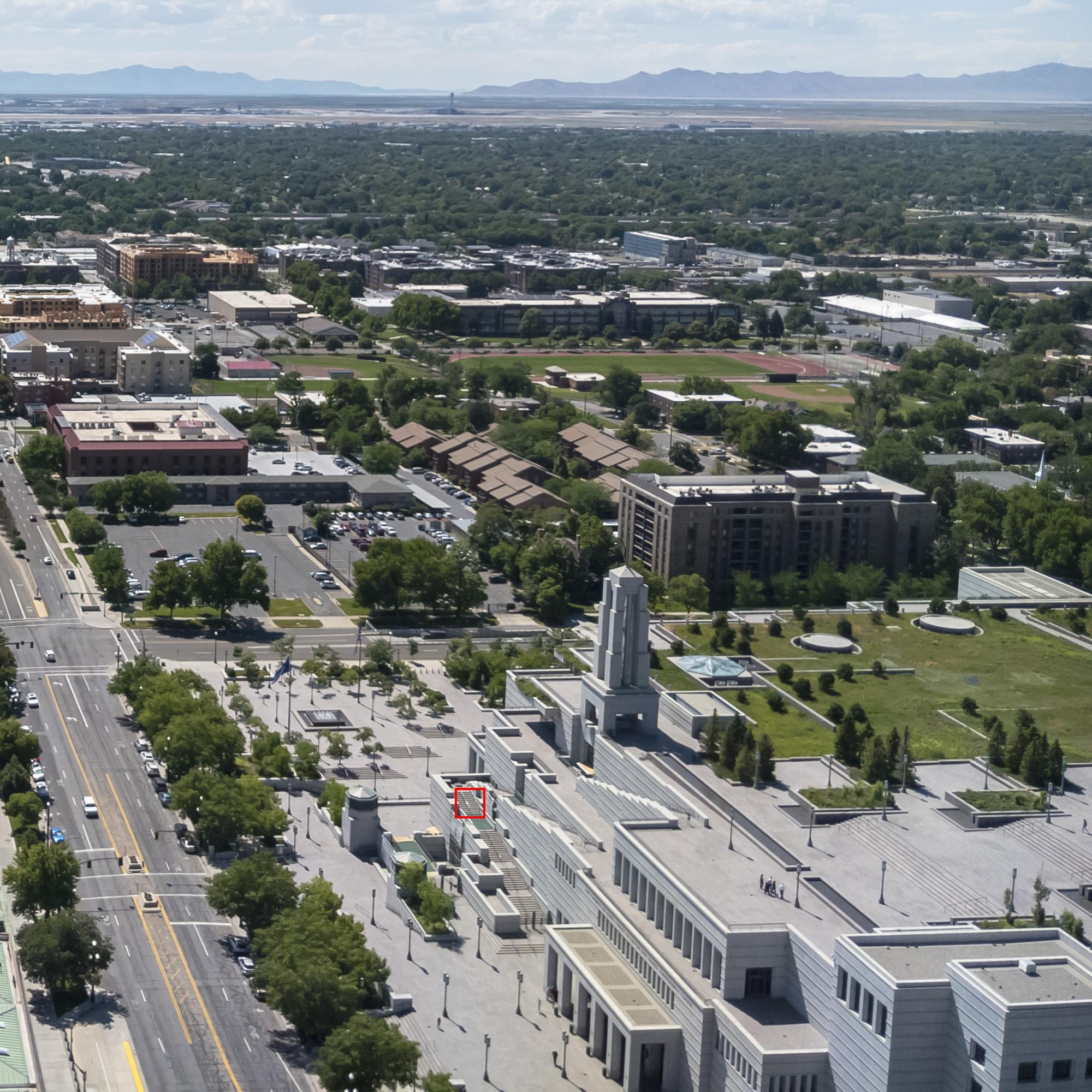}\\
Image captured by Google Pixel 2
\end{tabular}
\end{adjustbox}
\scriptsize
\begin{adjustbox}{valign=t}
\begin{tabular}{ccc}
\includegraphics[width=0.172\linewidth]{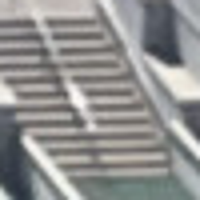}      &
\includegraphics[width=0.172\linewidth]{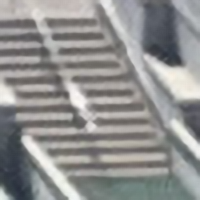} 	      &
\includegraphics[width=0.172\linewidth]{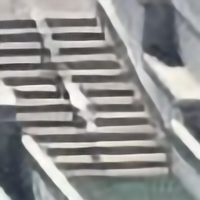}          \\
Bicubic  &RCAN + BD     &RCAN + MD                                        \\

\includegraphics[width=0.172\linewidth]{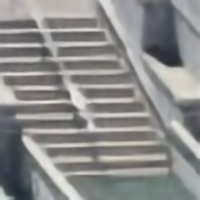}  	      &
\includegraphics[width=0.172\linewidth]{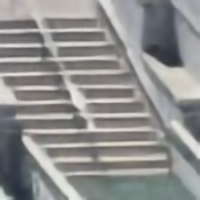}            &
\includegraphics[width=0.172\linewidth]{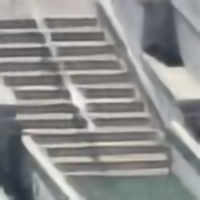}  	      \\ 
RCAN + RealSR    &KPN ($k$ = 19) + RealSR     &LP-KPN + RealSR            \\
\end{tabular}
\end{adjustbox}   
\end{tabular}
\caption{SISR results ($\times4$) of real-world images outside our dataset. Images are captured by Sony a7II, iPhone X and Google Pixel 2.}
\label{sub:cross}
\end{figure*}
\subsection{More super-resolved results on images outside our dataset}
In this subsection, we provide more super-resolved results on images outside our dataset, including images taken by one Sony a7II DSLR camera and two mobile cameras (\ie, iPhone X and Google Pixel 2).
The visual examples are shown in Fig. \ref{sub:cross}.
%

%---------------------------------------------------------------------------------------------
{
\bibliographystyle{ieee}
\bibliography{egbib}
}

\end{document}